\documentclass{article}
\usepackage[T1]{fontenc}
\usepackage[utf8]{inputenc}
\usepackage{arxiv}
\usepackage{amsmath,amssymb,amsthm}
\usepackage{graphicx,xcolor}
\usepackage[authoryear]{natbib}
\setcitestyle{authoryear,open={[},close={]},citesep={;},aysep={},yysep={,},notesep={, }}
\usepackage[colorlinks=true,linkcolor=blue,citecolor=blue,urlcolor=blue]{hyperref}
\usepackage{defs}
\usepackage[capitalise]{cleveref}
\usepackage{subcaption,multirow,booktabs,array,float}
\usepackage{fvextra}
\input{code-style.tex}
\DefineVerbatimEnvironment{CodeBlock}{Verbatim}{
  commandchars=\\\{\},
  bgcolor=gray!5,
  frame=single,
  framesep=2mm,
  fontsize=\small,
  breaklines=true
}
\VerbatimPygments{\PYG}{\PYG}
\renewcommand{\shorttitle}{GPU-Enabled Large-Scale Optimization Using RandNLA}
\renewcommand{\headeright}{Preprint}
\renewcommand{\undertitle}{Preprint}
\title{GPU-Enabled Large-Scale Optimization Using Randomized Linear Algebra}
\author{\normalfont Pratik Rathore\thanks{Both authors contributed equally to this work.} \quad Zachary Frangella\textsuperscript{*} \quad Parth Nobel \\
\normalfont Xuning Hu \quad Madeleine Udell \\[1ex]
\normalfont Stanford University \\
\normalfont\small \href{mailto:pratikr@alumni.stanford.edu}{\texttt{pratikr@alumni.stanford.edu}} \quad \href{mailto:zfrangella@alumni.stanford.edu}{\texttt{zfrangella@alumni.stanford.edu}} \quad \href{mailto:ptnobel@alumni.stanford.edu}{\texttt{ptnobel@alumni.stanford.edu}} \\
\normalfont\small \href{mailto:xuningh@stanford.edu}{\texttt{xuningh@stanford.edu}} \quad \href{mailto:udell@stanford.edu}{\texttt{udell@stanford.edu}}}
\date{}
\hypersetup{pdftitle={GPU-Enabled Large-Scale Optimization Using Randomized Linear Algebra},pdfauthor={Pratik Rathore, Zachary Frangella, Parth Nobel, Xuning Hu, Madeleine Udell}}
\begin{document}
\maketitle
\begin{abstract}
This paper introduces \rlaopt{}, a PyTorch-based package for large-scale optimization and scientific computing using randomized numerical linear algebra (RandNLA).
    Despite substantial progress in RandNLA-based algorithms, few implementations combine GPU acceleration with a simple interface for specifying optimization problems.
    \rlaopt{} addresses this gap by providing GPU-enabled solvers for positive-definite linear systems and convex empirical risk minimization with constraints and regularizers.
    These solvers use RandNLA to accelerate conjugate gradient (\nys{}PCG), operator splitting (NysADMM), and stochastic gradient methods (SAPPHIRE).
    Moreover, \rlaopt{} includes a modeling language that lets users specify problems using natural mathematical syntax.
    \rlaopt{} automatically checks compatibility with the selected solver and performs the required problem decomposition.
    The solvers also support differentiation through their iterations, enabling applications such as hyperparameter tuning.
    Experiments on ridge regression, bounded multinomial logistic regression, and bounded elastic net identify when randomized preconditioning improves performance and demonstrate substantial speedups from GPU execution.
    The package is open-source under an Apache license, with source code at \href{https://github.com/udellgroup/rlaopt}{https://github.com/udellgroup/rlaopt} and version 0.1.0 available on \href{https://pypi.org/project/rlaopt/0.1.0/}{PyPI}.
\end{abstract}
\section{Introduction}
\label{sec:rlaopt-introduction}
Large-scale optimization lies at the heart of modern machine learning and scientific computing.
Datasets routinely contain millions of samples and features, giving rise to optimization problems whose sheer scale demands efficient algorithms and hardware-aware implementations.
Over the past two decades, randomized numerical linear algebra (RandNLA) has emerged as a powerful toolkit for addressing these challenges, producing algorithms that exploit low-rank structure and stochastic approximations to dramatically reduce computational costs \citep{halko2011finding, mahoney2011randomized, woodruff2014sketching, martinsson2020randomized}.
However, a significant gap remains between the strong algorithmic foundations of RandNLA and the software available to practitioners:
existing implementations of RandNLA-based methods lack the ability to leverage modern parallel hardware such as GPUs, and do not provide a simple, user-friendly syntax for modeling optimization problems.
Consequently, practitioners who wish to apply RandNLA-based algorithms to large-scale problems must piece together ad hoc implementations, limiting both accessibility and performance.

We now describe two important problem classes where RandNLA-based algorithms offer significant advantages, and where the lack of high-quality implementations is particularly acute.

\paragraph{Large-scale linear systems.}
Dense linear systems of the form $(A + \lambda I) x = b$ arise in kernel ridge regression \citep{scholkopf2002learning}, Gaussian process inference \citep{rasmussen2005gaussian, gardner2018gpytorch}, and other settings throughout machine learning and scientific computing.
Direct methods such as Cholesky decomposition require $\bigO(n^3)$ computation and $\bigO(n^2)$ storage, limiting them to problems with $n \sim 10^4$.
Iterative methods such as conjugate gradient scale more favorably, with per-iteration complexity $\bigO(n^2)$, but converge slowly when the problem is ill-conditioned---a common occurrence in practice.
RandNLA-based preconditioning, in particular the randomized \nys{} preconditioner \citep{frangella2023randomized}, dramatically accelerates convergence by constructing high-quality low-rank approximations of the kernel matrix at modest cost.
The resulting preconditioned conjugate gradient (\nys{}PCG) method combines the scalability of iterative methods with robustness to ill-conditioning.

\paragraph{Large-scale optimization.}
Classical optimization methods, such as interior-point methods, produce high-accuracy solutions but rely on expensive matrix factorizations that limit their applicability to large-scale problems \citep{odonoghue2016conic, stellato2020osqp, applegate2021practical}.
On the other end of the spectrum, first-order methods such as stochastic gradient descent (SGD) and its variants \citep{robbins1951stochastic, johnson2013accelerating, defazio2014saga, allenzhu2018katyusha} scale to massive datasets but suffer from slow convergence on ill-conditioned problems and sensitivity to hyperparameters such as the learning rate \citep{nemirovski2009robust}.
Operator splitting frameworks such as the alternating direction method of multipliers (ADMM) \citep{boyd2011distributed} naturally handle constraints and nonsmooth regularizers, but also require the solution of subproblems that can be ill-conditioned.
RandNLA techniques integrate naturally with both stochastic gradient methods and operator splitting:
\nys{}-based preconditioning accelerates ADMM by improving the conditioning of linear system subproblems \citep{zhao2022nysadmm, diamandis2026genios}, and randomized curvature estimates yield preconditioned stochastic gradient methods with reliable default hyperparameters and fast convergence \citep{frangella2024sketchysgd, frangella2024promise, sun2025sapphire}.
Crucially, the dominant operations in these RandNLA-enhanced methods are matrix-matrix and matrix-vector products, which are highly amenable to GPU acceleration.

To address the gap between RandNLA algorithms and practical software, we develop \rlaopt{}, a PyTorch-based package for large-scale optimization and scientific computing.
\rlaopt{} includes GPU-enabled implementations of \nys{}PCG for large-scale positive-definite linear systems, NysADMM \citep{zhao2022nysadmm} for constrained convex optimization, and SAPPHIRE \citep{frangella2024promise, sun2025sapphire} for empirical risk minimization.
To make these solvers accessible, we create a flexible modeling language inspired by disciplined convex programming \citep{grant2006disciplined} and CVXPY \citep{diamond2016cvxpy} that allows users to specify optimization problems using natural mathematical syntax.
\rlaopt{} also supports differentiating through the solver, making it suitable for modern machine learning pipelines that require end-to-end gradient computation.
Our implementation is open-sourced under an Apache license and is available at \href{https://github.com/udellgroup/rlaopt}{https://github.com/udellgroup/rlaopt}.
Documentation can be found at \href{https://rlaopt.readthedocs.io}{https://rlaopt.readthedocs.io}, and version 0.1.0 is available on \href{https://pypi.org/project/rlaopt/0.1.0/}{PyPI}.

\subsection{Contributions}
Our contributions are as follows:
\begin{enumerate}
    \item We develop \rlaopt{}, an open-source, PyTorch-based software package for large-scale optimization using randomized linear algebra. \rlaopt{} provides a shared solver interface for CPU and GPU execution.
    \item We create a modeling language, inspired by disciplined convex programming \citep{grant2006disciplined} and CVXPY \citep{diamond2016cvxpy}, that lets users compose losses, regularizers, and constraints using natural mathematical syntax.
          \rlaopt{} checks compatibility with the user's selected solver and automatically introduces auxiliary variables and linear constraints for ADMM splitting.
    \item We implement a suite of RandNLA-based algorithms within \rlaopt{}, including \nys{}PCG for large-scale positive-definite linear systems, NysADMM for constrained convex optimization via operator splitting, and SAPPHIRE for preconditioned, stochastic variance-reduced optimization.
    \item We evaluate the methods in \rlaopt{} against state-of-the-art competitor methods for large-scale ridge regression, bounded multinomial logistic regression, and bounded elastic net regression.
    The experiments characterize the types of problems for which the RandNLA-based solvers in \rlaopt{} outcompete state-of-the-art competitors, and identify regimes where GPU provides significant speedups over CPU.
    \item We show that \rlaopt{} supports differentiating through the solver, enabling applications such as hyperparameter tuning and end-to-end learning within modern machine learning pipelines.
\end{enumerate}

\subsection{Roadmap}
\cref{sec:rlaopt-problem-classes} formally defines the problem classes addressed by \rlaopt{}.
\cref{sec:rlaopt-related-work} surveys related work organized by problem class.
\cref{sec:rlaopt-modeling-language} describes the modeling language and demonstrates its flexibility through examples.
\cref{sec:rlaopt-automatic-detection} explains how \rlaopt{} automatically detects problem structure, checks compatibility with the user-selected solver, and performs the required decomposition.
\cref{sec:rlaopt-experiments} presents performance benchmarks comparing CPU and GPU implementations across problem classes.
\cref{sec:rlaopt-conclusion} concludes the paper.
\section{Problem Classes}
\label{sec:rlaopt-problem-classes}
In this section, we formally define the problem classes addressed by \rlaopt{}.
We consider two broad classes: positive definite linear systems (\cref{subsec:rlaopt-pd-linear-systems}) and empirical risk minimization with constraints and regularizers (\cref{subsec:rlaopt-erm}).

\subsection{Positive-Definite Linear Systems}
\label{subsec:rlaopt-pd-linear-systems}
The first problem class consists of symmetric positive-definite (pd) linear systems of the form
\begin{equation}
    \label{eq:rlaopt-pd-linear-system}
    A x = b,
\end{equation}
where $A \in \R^{n \times n}$ is symmetric and pd, $b \in \R^n$, and $x \in \R^n$ is the unknown.
A common special case is the regularized linear system
\begin{equation}
    \label{eq:rlaopt-regularized-linear-system}
    (K + \lambda I) x = b,
\end{equation}
where $K \in \R^{n \times n}$ is symmetric positive-semidefinite (psd) and $\lambda > 0$ is a regularization parameter.

Linear systems of this form arise frequently in machine learning and scientific computing.
A prominent example is kernel ridge regression (KRR), in which one solves
\begin{equation}
    \label{eq:rlaopt-krr}
    \underset{w \in \R^n}{\text{minimize}} \quad \frac{1}{2} \| K w - y \|^2 + \frac{\lambda}{2} \| w \|_K^2,
\end{equation}
where $K \in \R^{n \times n}$ is a kernel matrix with entries $K_{ij} = k(x_i, x_j)$ for a kernel function $k$, and $y \in \R^n$ is the target vector.
The optimality conditions of \eqref{eq:rlaopt-krr} yield the linear system $(K + \lambda I) w^\star = y$.
Linear systems of the form \eqref{eq:rlaopt-regularized-linear-system} also arise in Gaussian process (GP) inference \citep{rasmussen2005gaussian}, where computing the posterior mean and variance requires solving dense linear systems involving a kernel matrix.

Direct methods such as Cholesky decomposition solve \eqref{eq:rlaopt-pd-linear-system} in $\bigO(n^3)$ time with $\bigO(n^2)$ storage, limiting them to problems with $n \lesssim 10^4$.
Iterative methods such as conjugate gradient (CG) scale more favorably, with per-iteration complexity $\bigO(n^2)$, but converge slowly when $A$ is ill-conditioned.
Ill-conditioning is common in practice: kernel matrices in machine learning often have rapidly decaying spectra, leading to large condition numbers \citep{caponnetto2007optimal, bach2013sharp, tu2016large, ma2017diving, belkin2018approximation}.
\rlaopt{} addresses this challenge by providing \nys{}PCG, which uses a randomized \nys{} preconditioner \citep{frangella2023randomized} to dramatically accelerate the convergence of CG on ill-conditioned systems.

\subsection{Empirical Risk Minimization with Constraints and Regularizers}
\label{subsec:rlaopt-erm}
The second problem class is composite convex optimization of the form
\begin{equation}
    \label{eq:rlaopt-composite-convex}
    \underset{x \in \R^n}{\text{minimize}} \quad f(x) + \sum_{i=1}^{k} g_i(A_i x - b_i),
\end{equation}
where $f \colon \R^n \to \R$ is smooth and convex, each $g_i \colon \R^{m_i} \to \R \cup \{+\infty\}$ is closed, convex, and proxable (i.e., its proximal operator can be evaluated efficiently), $A_i \in \R^{m_i \times n}$, and $b_i \in \R^{m_i}$.
This formulation is flexible enough to encode a wide range of machine learning problems, including empirical risk minimization (ERM) with constraints and regularizers.

In a typical ERM setting, the smooth component $f$ takes the form
\begin{equation}
    \label{eq:rlaopt-erm}
    f(x) = \frac{1}{N} \sum_{j=1}^{N} \ell_j(a_j^T x),
\end{equation}
where $\{(a_j, y_j)\}_{j=1}^N$ is a training set with $a_j \in \R^n$ and $y_j \in \R$, and $\ell_j$ is a loss function.
The nonsmooth terms $g_i$ encode regularizers and constraints such as $\ell_1$ regularization, box constraints, elastic net penalties, and indicator functions of convex sets.

Several concrete problems fit naturally into this framework:
\begin{itemize}
    \item \textbf{$\ell_1$-regularized logistic regression:} $f$ is the average logistic loss and $g(x) = \mu \| x \|_1$ for some regularization weight $\mu > 0$.
    \item \textbf{Bounded elastic net:} $f(w) = \frac{1}{2N} \| X w - y \|_2^2$, with $g_1(w) = \lambda_1 \| w \|_1 + \frac{\lambda_2}{2} \| w \|_2^2$ and $g_2$ encoding box constraints $0 \leq w \leq 1$.
    \item \textbf{Constrained multinomial regression:} $f$ is the average cross-entropy loss for multiclass classification, with $g$ encoding box constraints on the regression coefficients.
\end{itemize}

\rlaopt{} provides several solvers for problems of the form \eqref{eq:rlaopt-composite-convex}.
For problems where the full gradient of $f$ is available, \rlaopt{} supports NysADMM \citep{zhao2022nysadmm} which combines ADMM with \nys{}-based preconditioning to accelerate linear system subproblems that arise in ADMM.
For large-scale ERM problems where computing a full gradient of $f$ at each iteration is too expensive, \rlaopt{} supports SAPPHIRE \citep{frangella2024promise, sun2025sapphire}, a family of preconditioned stochastic variance-reduced methods that uses randomized curvature estimates to achieve fast convergence with reliable default hyperparameters.
\rlaopt{} also supports (accelerated) proximal gradient for problems where preconditioning is not needed.

\section{Related Work}
\label{sec:rlaopt-related-work}
We survey related work organized by the problem classes introduced in \cref{sec:rlaopt-problem-classes}.
For each class, we describe existing algorithmic approaches and software, and discuss how \rlaopt{} relates to and improves upon them.

\subsection{Large-Scale Linear Systems}

\paragraph{Direct methods.}
Direct methods such as Cholesky decomposition are the standard approach for solving dense pd linear systems \citep{golub2013matrix}.
While they produce high-accuracy solutions, their $\bigO(n^3)$ computational cost and $\bigO(n^2)$ storage requirements render them impractical for problems with $n \gtrsim 10^4$.
\rlaopt{} targets the regime where direct methods are too expensive, providing iterative solvers that scale to much larger problem sizes.

\paragraph{Iterative methods and preconditioning.}
Conjugate gradient (CG) is the method of choice for large-scale pd linear systems, with per-iteration complexity $\bigO(n^2)$ for dense systems.
However, CG converges slowly when the system is ill-conditioned, and effective preconditioners are essential for practical performance.
Randomized \nys{} preconditioning \citep{frangella2023randomized} constructs a high-quality low-rank approximation of the coefficient matrix using sketching techniques from RandNLA \citep{halko2011finding, martinsson2020randomized}.
The resulting \nys{}PCG method converges rapidly even on ill-conditioned problems, and its dominant cost---matrix-matrix products for constructing the preconditioner---is highly GPU-amenable.
\rlaopt{} provides a GPU-enabled implementation of \nys{}PCG; \cref{sec:rlaopt-experiments} evaluates its performance across spectral decay rates and regularization levels.

\paragraph{Kernel methods.}
FALKON \citep{rudi2017falkon} is a widely used solver for inducing points kernel ridge regression that uses \nys{} preconditioning with CG.
Gaussian process inference packages such as GPyTorch \citep{gardner2018gpytorch} also rely on PCG-based linear solvers.
\rlaopt{} differs from these tools by targeting general pd linear systems rather than a specific application and providing a modeling language for specifying the problem.

\subsection{Optimization Solvers}

\paragraph{Interior-point methods.}
Interior-point methods (IPMs) are the gold standard for small-to-moderate-scale convex optimization, producing high-accuracy solutions with polynomial-time guarantees \citep{yurii1994interiorpoint}.
Software implementations such as Clarabel \citep{chen2024clarabel} provide reliable IPM solvers, and recent work has extended these to GPU \citep{chen2025cuclarabel}.
However, IPMs rely on matrix factorizations whose cost grows cubically with problem size, limiting their applicability to large-scale machine learning problems.
\rlaopt{} takes a complementary approach, using first-order and operator splitting methods enhanced with RandNLA to handle problems at scales where IPMs are impractical.

\paragraph{Operator splitting methods.}
The alternating direction method of multipliers (ADMM) \citep{gabay1976dual, boyd2011distributed} and related operator splitting methods \citep{ryu2022largescale} decompose composite optimization problems into simpler subproblems that can be solved independently.
ADMM is the basis for several widely used solvers, including OSQP \citep{stellato2020osqp} for quadratic programs and SCS \citep{odonoghue2016conic} for conic programs.
A key bottleneck in ADMM is solving a subproblem corresponding to the primal update.
NysADMM \citep{zhao2022nysadmm} accelerates inexact ADMM by solving the primal subproblem with \nys{}PCG, using a randomized Nyström preconditioner to handle ill-conditioning; for non-quadratic smooth losses, the subproblem is formed from a second-order approximation of the smooth term.
GeNIOS \citep{diamandis2026genios} builds on this recipe with adaptive penalty parameter updates, periodic preconditioner refresh, an adaptive inexact PCG tolerance, ADMM over-relaxation, and primal/dual infeasibility detection, with an open-source Julia implementation.
\rlaopt{}'s NysADMM solver inherits these engineering choices from GeNIOS (with the exception of infeasibility detection), adapted to a PyTorch/GPU setting.
A related line of work specializes operator splitting to linear programming: PDLP \citep{applegate2021practical} applies the primal-dual hybrid gradient method to a saddle-point formulation of LP with diagonal preconditioning, adaptive step sizes, and adaptive restarts, and cuPDLP \citep{lu2025cupdlp} ports this approach to GPUs.
These solvers share \rlaopt{}'s philosophy of scaling on GPU through matrix-vector products, but target LP specifically rather than the broader composite convex setting.

\paragraph{Proximal gradient methods.}
Proximal gradient methods \citep{parikh2014proximal} and their accelerated variants are a natural fit for composite problems of the form \eqref{eq:rlaopt-composite-convex} when the proximal operator of each $g_i$ is cheap to evaluate.
These methods have per-iteration costs dominated by gradient evaluations of the smooth component $f$, but converge slowly on ill-conditioned problems.
\rlaopt{} implements proximal gradient with optional \nys{}-based preconditioning, which improves convergence on ill-conditioned objectives while preserving the simplicity of the proximal gradient framework.

\paragraph{Stochastic gradient methods.}
When the smooth component $f$ has a finite-sum structure, stochastic gradient methods such as SGD \citep{robbins1951stochastic}, Adam \citep{kingma2014adam}, SVRG \citep{johnson2013accelerating}, SAGA \citep{defazio2014saga}, and Katyusha \citep{allenzhu2018katyusha, kovalev2020lkatyusha} achieve low per-iteration costs by operating on mini-batches.
However, these methods are sensitive to hyperparameters and converge slowly on ill-conditioned problems \citep{nemirovski2009robust}.
The PROMISE framework \citep{frangella2024promise} and SketchySGD \citep{frangella2024sketchysgd} address these limitations by using randomized curvature estimates as preconditioners, yielding methods with reliable default hyperparameters and fast convergence on ill-conditioned problems.
SAPPHIRE \citep{sun2025sapphire} generalizes this line of work to proximal settings, supporting nonsmooth regularizers and constraints within the preconditioned stochastic gradient framework.
\rlaopt{} provides GPU-enabled implementations of SAPPHIRE methods; \cref{sec:rlaopt-experiments} compares their CPU and GPU performance with deterministic baselines on bounded multinomial logistic regression.

\paragraph{Differentiable optimization.}
Several frameworks enable differentiating through the solution of optimization problems.
cvxpylayers \citep{agrawal2019differentiable} embeds parametrized convex programs specified in CVXPY as differentiable layers, using implicit differentiation through the KKT conditions of the conic reformulation.
This approach supports a broad class of disciplined convex programs. Its computational cost depends on the conic reformulation, the solver, and the linear algebra used for differentiation; the diffcp backend supports iterative linear solves for derivative evaluation.
JAXopt \citep{blondel2022efficient} provides a modular implicit differentiation framework in JAX with a wide range of solvers, including proximal gradient, L-BFGS, and OSQP.
MPAX \citep{lu2024mpax} is a JAX-based first-order solver for large-scale linear and quadratic programs that supports differentiation through unrolled solver iterations.
These frameworks are complementary to \rlaopt{}: they cover different problem classes or ecosystems, while \rlaopt{} contributes differentiable, RandNLA-based solvers in PyTorch that are specifically designed for large-scale, ill-conditioned problems.

\section{Modeling Language}
\label{sec:rlaopt-modeling-language}
A central design goal of \rlaopt{} is to provide a simple, expressive interface for specifying optimization problems.
To this end, we develop a modeling language inspired by disciplined convex programming \citep{grant2006disciplined} and CVXPY \citep{diamond2016cvxpy} that lets users construct objectives from composable building blocks using natural mathematical syntax.
In this section, we describe the core abstractions of the modeling language and demonstrate its flexibility through examples.

\subsection{Core Abstractions}
The \rlaopt{} modeling language is built around three core abstractions: \emph{variables}, \emph{atoms}, and \emph{solvers}.

\paragraph{Variables.}
A \texttt{Variable} represents an unknown quantity to be optimized.
Variables are created by specifying their shape and, optionally, a name and a device (CPU or GPU):
\begin{CodeBlock}
\PYG{n}{w} \PYG{o}{=} \PYG{n}{Variable}\PYG{p}{(}\PYG{p}{(}\PYG{n}{n}\PYG{p}{,}\PYG{p}{)}\PYG{p}{,} \PYG{n}{name}\PYG{o}{=}\PYG{l+s+s2}{\PYGZdq{}}\PYG{l+s+s2}{w}\PYG{l+s+s2}{\PYGZdq{}}\PYG{p}{)}
\PYG{n}{beta} \PYG{o}{=} \PYG{n}{Variable}\PYG{p}{(}\PYG{p}{(}\PYG{n}{p}\PYG{p}{,} \PYG{n}{K}\PYG{p}{)}\PYG{p}{,} \PYG{n}{name}\PYG{o}{=}\PYG{l+s+s2}{\PYGZdq{}}\PYG{l+s+s2}{beta}\PYG{l+s+s2}{\PYGZdq{}}\PYG{p}{,} \PYG{n}{device}\PYG{o}{=}\PYG{l+s+s2}{\PYGZdq{}}\PYG{l+s+s2}{cuda}\PYG{l+s+s2}{\PYGZdq{}}\PYG{p}{)}
\end{CodeBlock}

Variables can appear in mathematical expressions involving matrix multiplication, addition, and subtraction, using standard Python operators.
For instance, \texttt{X @ w + b - y} represents the affine expression $Xw + b - y$.

\paragraph{Atoms.}
An \emph{atom} is a function with known mathematical properties (e.g., smoothness, or the availability of a proximal operator) that serves as a building block for constructing objectives.
\rlaopt{} provides atoms for common losses, regularizers, and constraints encountered in machine learning and scientific computing.
\cref{tab:rlaopt-atoms} summarizes the available atoms.

\begin{table}[t]
    \centering
    \small
    \setlength{\tabcolsep}{3pt}
    \begin{tabular}{@{}lll@{}}
        \toprule
        \textbf{Category} & \textbf{Atom}                                    & \textbf{Mathematical form}                              \\
        \midrule
        \multirow{2}{*}{General}
                          & \texttt{SumSquares(expr)}                        & $\| \text{expr} \|_2^2$                                 \\
                          & \texttt{QuadForm(expr, Q)}                       & $\text{expr}^T Q\,\text{expr}$                           \\
        \midrule
        \multirow{9}{*}{\shortstack[l]{Linear model\\losses}}
                          & \texttt{LinearRegression(w, loader)}             & $\frac{1}{N}\sum_{j} (y_j - z_j)^2$              \\
                          & \texttt{LogisticRegression(w, loader)}           & $\frac{1}{N}\sum_{j} [\log(1 + e^{z_j})-y_jz_j]$     \\
                          & \texttt{MultinomialRegression(w, loader)}        & $-\frac{1}{N}\sum_{j} \log(\text{softmax}(z_j)_{c_j})$ \\
                          & \texttt{PoissonRegression(w, loader)}            & Poisson negative log-likelihood                         \\
                          & \texttt{GammaRegression(w, loader)}              & Gamma negative log-likelihood                           \\
                          & \texttt{InverseGaussianRegression(w, loader)}    & Inverse Gaussian negative log-likelihood                \\
                          & \shortstack[l]{\texttt{CompoundPoissonGammaRegression}\\\texttt{(w, loader, power=$q$)}} & Tweedie loss, $1<q<2$ \\
                          & \texttt{HuberRegression(w, loader, delta=$\delta$)} & $\frac{1}{N}\sum_j h_\delta(y_j-z_j)$ \\
        \midrule
        \multirow{5}{*}{Regularizers}
                          & \texttt{L1Norm(w, $\lambda$)}                    & $\lambda \| w \|_1$                                     \\
                          & \texttt{L2Norm(w, $\lambda$)}                    & $\lambda \| w \|_2$                                     \\
                          & \texttt{LInfNorm(w, $\lambda$)}                  & $\lambda \| w \|_\infty$                                \\
                          & \texttt{NucNorm(W, $\lambda$)}                   & $\lambda \| W \|_*$ (nuclear norm)                      \\
                          & \texttt{ElasticNet(w, $\lambda_1$, $\lambda_2$)} & $\lambda_1 \| w \|_1 + \frac{\lambda_2}{2} \| w \|_2^2$ \\
        \midrule
        \multirow{8}{*}{Constraints}
                          & \texttt{Box(w, lower, upper)}                    & $\I[\text{lower} \leq w \leq \text{upper}]$             \\
                          & \texttt{NonNegative(w)}                          & $\I[w \geq 0]$                                          \\
                          & \texttt{Halfspace(w, c, upper)}                  & $\I[c^T w \leq \text{upper}]$                        \\
                          & \texttt{LinearEquality(w, A, b)}                 & $\I[Aw = b]$                                            \\
                          & \texttt{Polyhedron(w, A, b, C, l, u)}            & $\I[Aw = b,\; l \leq Cw \leq u]$                        \\
                          & \texttt{L1NormBall(w, r)}                        & $\I[\| w \|_1 \leq r]$                                  \\
                          & \texttt{L2NormBall(w, r)}                        & $\I[\| w \|_2 \leq r]$                                  \\
                          & \texttt{LInfNormBall(w, r)}                      & $\I[\| w \|_\infty \leq r]$                             \\
        \bottomrule
    \end{tabular}
    \caption[Atoms available in \rlaopt{}.]{Atoms available in \rlaopt{}. Linear model losses operate on data provided via a \texttt{DataLoader}. $\I[\cdot]$ denotes the indicator function of the given constraint set (zero when satisfied, $+\infty$ otherwise). $\|\cdot\|_*$ denotes the nuclear norm (sum of singular values).}
    \label{tab:rlaopt-atoms}
\end{table}

\paragraph{Composing objectives.}
Objectives are constructed by combining atoms with the \texttt{+} operator, mirroring the mathematical structure of the problem.
Scalar multiplication via \texttt{*} is also supported.
For example, the bounded elastic net problem
\begin{equation*}
    \underset{w, b}{\text{minimize}} \quad \frac{1}{2N} \| X w + b - y \|_2^2 + \lambda_1 \| w \|_1 + \frac{\lambda_2}{2} \| w \|_2^2 \quad \text{subject to} \quad 0 \leq w \leq 1
\end{equation*}
is specified as:
\begin{CodeBlock}
\PYG{n}{w} \PYG{o}{=} \PYG{n}{Variable}\PYG{p}{(}\PYG{p}{(}\PYG{n}{X}\PYG{o}{.}\PYG{n}{shape}\PYG{p}{[}\PYG{l+m+mi}{1}\PYG{p}{]}\PYG{p}{,}\PYG{p}{)}\PYG{p}{)}
\PYG{n}{b} \PYG{o}{=} \PYG{n}{Variable}\PYG{p}{(}\PYG{p}{(}\PYG{l+m+mi}{1}\PYG{p}{,}\PYG{p}{)}\PYG{p}{)}
\PYG{n}{obj} \PYG{o}{=} \PYG{n}{SumSquares}\PYG{p}{(}\PYG{n}{X} \PYG{o}{@} \PYG{n}{w} \PYG{o}{+} \PYG{n}{b} \PYG{o}{\PYGZhy{}} \PYG{n}{y}\PYG{p}{)} \PYG{o}{*} \PYG{p}{(}\PYG{l+m+mf}{0.5} \PYG{o}{/} \PYG{n}{N}\PYG{p}{)} \PYGZbs{}
    \PYG{o}{+} \PYG{n}{ElasticNet}\PYG{p}{(}\PYG{n}{w}\PYG{p}{,} \PYG{n}{lam1}\PYG{p}{,} \PYG{n}{lam2}\PYG{p}{)} \PYGZbs{}
    \PYG{o}{+} \PYG{n}{Box}\PYG{p}{(}\PYG{n}{w}\PYG{p}{,} \PYG{l+m+mf}{0.0}\PYG{p}{,} \PYG{l+m+mf}{1.0}\PYG{p}{)}
\end{CodeBlock}

The resulting objective \texttt{obj} automatically tracks the variables and atoms that compose it.

\paragraph{Solvers.}
Once an objective has been defined, the user selects a solver and its configuration.
\rlaopt{} provides a unified solver interface with two modes of operation: a \emph{stepped} mode for fine-grained control over the optimization loop, and a \emph{direct} mode for one-call solving.
A complete reference of all solver configuration parameters, termination criteria, and their defaults is provided in \cref{app:rlaopt-solver-configs}.
In stepped mode, the user initializes the solver state and calls \texttt{step} iteratively (note that the NysADMM solver is accessed via the \texttt{ADMM} class in the API):
\begin{CodeBlock}
\PYG{n}{solver} \PYG{o}{=} \PYG{n}{ADMM}\PYG{p}{(}\PYG{n}{obj}\PYG{p}{,} \PYG{n}{config}\PYG{o}{=}\PYG{n}{ADMMConfig}\PYG{p}{(}\PYG{p}{)}\PYG{p}{)}
\PYG{n}{variable\PYGZus{}values} \PYG{o}{=} \PYG{n}{obj}\PYG{o}{.}\PYG{n}{variable\PYGZus{}values}
\PYG{n}{state} \PYG{o}{=} \PYG{n}{solver}\PYG{o}{.}\PYG{n}{init\PYGZus{}state}\PYG{p}{(}\PYG{n}{variable\PYGZus{}values}\PYG{p}{)}
\PYG{k}{for} \PYG{n}{\PYGZus{}} \PYG{o+ow}{in} \PYG{n+nb}{range}\PYG{p}{(}\PYG{n}{num\PYGZus{}iters}\PYG{p}{)}\PYG{p}{:}
    \PYG{n}{variable\PYGZus{}values}\PYG{p}{,} \PYG{n}{state} \PYG{o}{=} \PYG{n}{solver}\PYG{o}{.}\PYG{n}{step}\PYG{p}{(}\PYG{n}{variable\PYGZus{}values}\PYG{p}{,} \PYG{n}{state}\PYG{p}{)}
\end{CodeBlock}

In direct mode, the user simply calls \texttt{solve}:
\begin{CodeBlock}
\PYG{n}{solver} \PYG{o}{=} \PYG{n}{ADMM}\PYG{p}{(}\PYG{n}{obj}\PYG{p}{,} \PYG{n}{config}\PYG{o}{=}\PYG{n}{ADMMConfig}\PYG{p}{(}\PYG{p}{)}\PYG{p}{)}
\PYG{n}{result} \PYG{o}{=} \PYG{n}{solver}\PYG{o}{.}\PYG{n}{solve}\PYG{p}{(}\PYG{p}{)}
\PYG{n}{x\PYGZus{}sol} \PYG{o}{=} \PYG{n}{result}\PYG{o}{.}\PYG{n}{variable\PYGZus{}values}
\end{CodeBlock}

Both modes return the solution as a dictionary mapping variable names to their optimal values.

\paragraph{Linear systems.}
For pd linear systems, \rlaopt{} provides a separate \texttt{LinSys} interface.
The user specifies a matrix, right-hand side, and regularization:
\begin{CodeBlock}
\PYG{n}{lin\PYGZus{}sys} \PYG{o}{=} \PYG{n}{LinSys}\PYG{p}{(}\PYG{n}{A}\PYG{p}{,} \PYG{n}{b}\PYG{p}{,} \PYG{n}{reg}\PYG{o}{=}\PYG{l+m+mf}{1e\PYGZhy{}2}\PYG{p}{)}
\end{CodeBlock}

The linear system is then solved with \nys{}PCG:
\begin{CodeBlock}
\PYG{n}{precond\PYGZus{}config} \PYG{o}{=} \PYG{n}{NystromConfig}\PYG{p}{(}\PYG{n}{rank\PYGZus{}init}\PYG{o}{=}\PYG{l+m+mi}{50}\PYG{p}{,} \PYG{n}{base\PYGZus{}damping}\PYG{o}{=}\PYG{l+m+mf}{1e\PYGZhy{}2}\PYG{p}{)}
\PYG{n}{pcg\PYGZus{}config} \PYG{o}{=} \PYG{n}{PCGConfig}\PYG{p}{(}\PYG{n}{preconditioner\PYGZus{}config}\PYG{o}{=}\PYG{n}{precond\PYGZus{}config}\PYG{p}{)}
\PYG{n}{solver} \PYG{o}{=} \PYG{n}{PCG}\PYG{p}{(}\PYG{n}{lin\PYGZus{}sys}\PYG{p}{,} \PYG{n}{config}\PYG{o}{=}\PYG{n}{pcg\PYGZus{}config}\PYG{p}{)}

\PYG{n}{params} \PYG{o}{=} \PYG{n}{lin\PYGZus{}sys}\PYG{o}{.}\PYG{n}{w}
\PYG{n}{state} \PYG{o}{=} \PYG{n}{solver}\PYG{o}{.}\PYG{n}{init\PYGZus{}state}\PYG{p}{(}\PYG{n}{params}\PYG{p}{)}
\PYG{k}{for} \PYG{n}{\PYGZus{}} \PYG{o+ow}{in} \PYG{n+nb}{range}\PYG{p}{(}\PYG{l+m+mi}{100}\PYG{p}{)}\PYG{p}{:}
    \PYG{n}{params}\PYG{p}{,} \PYG{n}{state} \PYG{o}{=} \PYG{n}{solver}\PYG{o}{.}\PYG{n}{step}\PYG{p}{(}\PYG{n}{params}\PYG{p}{,} \PYG{n}{state}\PYG{p}{)}
\end{CodeBlock}

\subsection{Differentiating Through the Solver}
A key feature of \rlaopt{} is its native support for differentiating through the optimization solver.
In modern machine learning pipelines, users are often interested in optimizing with respect to a parameter of the optimization problem itself---for example, tuning a regularization parameter to minimize validation error, or learning loss function parameters in an end-to-end pipeline \citep{agrawal2019differentiable, blondel2022efficient, lu2024mpax}.
Formally, consider a parametrized optimization problem $x^\star(\theta) = \text{argmin}_{x} \, \phi(x; \theta)$, where $\theta \in \R^d$ is an external parameter.
A common use case is hyperparameter tuning, in which one seeks to minimize an outer objective that depends on the solution of the inner problem:
\begin{equation}
    \label{eq:rlaopt-bilevel}
    \underset{\theta \in \R^d}{\text{minimize}} \quad \mathcal{L}(x^\star(\theta)),
\end{equation}
where $\mathcal{L}$ is a validation loss or other performance metric.
For example, one may wish to tune the regularization parameter $\mu$ in a lasso problem to minimize the prediction error on a held-out validation set:
\begin{equation}
    \label{eq:rlaopt-lasso-tuning}
    \underset{\mu > 0}{\text{minimize}} \quad \frac{1}{N_{\text{val}}} \| X_{\text{val}} \, x^\star(\mu) - y_{\text{val}} \|_2^2, \quad \text{where} \quad x^\star(\mu) = \underset{x}{\text{argmin}} ~ \frac{1}{N_{\text{train}}} \| X_{\text{train}} x - y_{\text{train}} \|_2^2 + \mu \| x \|_1.
\end{equation}

The solvers in \rlaopt{} are implemented in a functional manner and support automatic differentiation by backpropagation through the optimization trajectory.
By default, differentiation through the solver is disabled to avoid unnecessary overhead, but users can enable it by passing \texttt{detach=False} to the solver constructor.
For the lasso tuning problem \eqref{eq:rlaopt-lasso-tuning}, this is expressed as:
\begin{CodeBlock}
\PYG{n}{config} \PYG{o}{=} \PYG{n}{ProxGradConfig}\PYG{p}{(}
    \PYG{n}{eta}\PYG{o}{=}\PYG{n+nb}{float}\PYG{p}{(}\PYG{n}{N\PYGZus{}train} \PYG{o}{/} \PYG{p}{(}\PYG{l+m+mi}{2} \PYG{o}{*} \PYG{n}{torch}\PYG{o}{.}\PYG{n}{linalg}\PYG{o}{.}\PYG{n}{matrix\PYGZus{}norm}\PYG{p}{(}\PYG{n}{A\PYGZus{}train}\PYG{p}{,} \PYG{n+nb}{ord}\PYG{o}{=}\PYG{l+m+mi}{2}\PYG{p}{)}\PYG{o}{*}\PYG{o}{*}\PYG{l+m+mi}{2}\PYG{p}{)}\PYG{p}{)}\PYG{p}{,}
    \PYG{n}{use\PYGZus{}linesearch}\PYG{o}{=}\PYG{k+kc}{False}\PYG{p}{,} \PYG{n}{use\PYGZus{}acceleration}\PYG{o}{=}\PYG{k+kc}{False}\PYG{p}{)}

\PYG{k}{def} \PYG{n+nf}{outer\PYGZus{}objective}\PYG{p}{(}\PYG{n}{mu}\PYG{p}{)}\PYG{p}{:}
    \PYG{n}{x} \PYG{o}{=} \PYG{n}{Variable}\PYG{p}{(}\PYG{n}{torch}\PYG{o}{.}\PYG{n}{zeros}\PYG{p}{(}\PYG{n}{p}\PYG{p}{,} \PYG{n}{dtype}\PYG{o}{=}\PYG{n}{mu}\PYG{o}{.}\PYG{n}{dtype}\PYG{p}{)}\PYG{p}{,} \PYG{n}{name}\PYG{o}{=}\PYG{l+s+s2}{\PYGZdq{}}\PYG{l+s+s2}{x}\PYG{l+s+s2}{\PYGZdq{}}\PYG{p}{)}
    \PYG{n}{train\PYGZus{}obj} \PYG{o}{=} \PYG{n}{SumSquares}\PYG{p}{(}\PYG{n}{A\PYGZus{}train} \PYG{o}{@} \PYG{n}{x} \PYG{o}{\PYGZhy{}} \PYG{n}{b\PYGZus{}train}\PYG{p}{)} \PYG{o}{*} \PYG{p}{(}\PYG{l+m+mi}{1} \PYG{o}{/} \PYG{n}{N\PYGZus{}train}\PYG{p}{)} \PYGZbs{}
              \PYG{o}{+} \PYG{n}{L1Norm}\PYG{p}{(}\PYG{n}{x}\PYG{p}{,} \PYG{n}{scaling}\PYG{o}{=}\PYG{n}{mu}\PYG{p}{)}
    \PYG{n}{solver} \PYG{o}{=} \PYG{n}{ProxGrad}\PYG{p}{(}\PYG{n}{train\PYGZus{}obj}\PYG{p}{,} \PYG{n}{config}\PYG{p}{,} \PYG{n}{detach}\PYG{o}{=}\PYG{k+kc}{False}\PYG{p}{)}
    \PYG{n}{values} \PYG{o}{=} \PYG{n}{train\PYGZus{}obj}\PYG{o}{.}\PYG{n}{variable\PYGZus{}values}
    \PYG{n}{state} \PYG{o}{=} \PYG{n}{solver}\PYG{o}{.}\PYG{n}{init\PYGZus{}state}\PYG{p}{(}\PYG{n}{values}\PYG{p}{)}
    \PYG{k}{for} \PYG{n}{\PYGZus{}} \PYG{o+ow}{in} \PYG{n+nb}{range}\PYG{p}{(}\PYG{l+m+mi}{100}\PYG{p}{)}\PYG{p}{:}
        \PYG{n}{values}\PYG{p}{,} \PYG{n}{state} \PYG{o}{=} \PYG{n}{solver}\PYG{o}{.}\PYG{n}{step}\PYG{p}{(}\PYG{n}{values}\PYG{p}{,} \PYG{n}{state}\PYG{p}{)}
    \PYG{k}{return} \PYG{p}{(}\PYG{p}{(}\PYG{n}{A\PYGZus{}val} \PYG{o}{@} \PYG{n}{values}\PYG{p}{[}\PYG{l+s+s2}{\PYGZdq{}}\PYG{l+s+s2}{x}\PYG{l+s+s2}{\PYGZdq{}}\PYG{p}{]} \PYG{o}{\PYGZhy{}} \PYG{n}{b\PYGZus{}val}\PYG{p}{)}\PYG{o}{*}\PYG{o}{*}\PYG{l+m+mi}{2}\PYG{p}{)}\PYG{o}{.}\PYG{n}{mean}\PYG{p}{(}\PYG{p}{)}

\PYG{n}{mu} \PYG{o}{=} \PYG{n}{torch}\PYG{o}{.}\PYG{n}{tensor}\PYG{p}{(}\PYG{l+m+mf}{0.2}\PYG{p}{,} \PYG{n}{dtype}\PYG{o}{=}\PYG{n}{torch}\PYG{o}{.}\PYG{n}{float64}\PYG{p}{)}
\PYG{k}{for} \PYG{n}{\PYGZus{}} \PYG{o+ow}{in} \PYG{n+nb}{range}\PYG{p}{(}\PYG{l+m+mi}{40}\PYG{p}{)}\PYG{p}{:}
    \PYG{n}{grad}\PYG{p}{,} \PYG{n}{value} \PYG{o}{=} \PYG{n}{torch}\PYG{o}{.}\PYG{n}{func}\PYG{o}{.}\PYG{n}{grad\PYGZus{}and\PYGZus{}value}\PYG{p}{(}\PYG{n}{outer\PYGZus{}objective}\PYG{p}{)}\PYG{p}{(}\PYG{n}{mu}\PYG{p}{)}
    \PYG{n}{mu} \PYG{o}{=} \PYG{p}{(}\PYG{n}{mu} \PYG{o}{\PYGZhy{}} \PYG{l+m+mf}{0.05} \PYG{o}{*} \PYG{n}{grad}\PYG{p}{)}\PYG{o}{.}\PYG{n}{detach}\PYG{p}{(}\PYG{p}{)}
\end{CodeBlock}

PyTorch's automatic differentiation propagates gradients through the solver iterations, yielding the derivative of the validation loss with respect to $\mu$ without requiring the user to implement custom backward passes.

\subsection{Comparison to Disciplined Convex Programming and CVXPY}
The \rlaopt{} modeling language shares disciplined convex programming (DCP) and CVXPY's philosophy of composable atoms and natural mathematical syntax, but differs in how problems are prepared for a solver.
DCP provides rules for constructing convex problems, while CVXPY transforms these problems into standard forms (e.g., conic forms) accepted by backend solvers \citep{grant2006disciplined,diamond2016cvxpy}.
However, conic reformulations can significantly increase the problem size and obscure the structure of the original problem.
In contrast, \rlaopt{} preserves the composite structure $f(x) + \sum_i g_i(A_i x - b_i)$ and works directly with gradient and proximal oracles, avoiding unnecessary reformulation.
This is particularly advantageous for machine learning problems such as logistic regression, where conic reformulation introduces many additional variables \citep{diamandis2026genios}.

\section{Automatic Detection of Problem Structure}
\label{sec:rlaopt-automatic-detection}
A key feature of \rlaopt{} is its ability to automatically detect the structure of a user-specified optimization problem and decompose it into a form amenable to the chosen solver.
This section describes how \rlaopt{} performs this decomposition for proximal gradient methods and ADMM-based methods.

\subsection{Smooth-Nonsmooth Decomposition}
Recall from \cref{sec:rlaopt-problem-classes} that \rlaopt{} handles composite convex optimization problems of the form
\begin{equation}
    \label{eq:rlaopt-composite-recall}
    \underset{x \in \R^n}{\text{minimize}} \quad f(x) + \sum_{i=1}^{k} g_i(A_i x - b_i),
\end{equation}
where $f$ is smooth and convex, and each $g_i$ is closed, convex, and proxable.
When a user constructs an objective by composing atoms (as described in \cref{sec:rlaopt-modeling-language}), \rlaopt{} must determine which terms constitute the smooth component $f$ and which terms constitute the nonsmooth components $g_i$, along with their associated linear operators $A_i$ and offsets $b_i$.

Every atom in \rlaopt{} declares two key properties: whether it is \emph{smooth} (i.e., differentiable everywhere) and whether it is \emph{proxable} (i.e., its proximal operator can be evaluated efficiently).
For example, \texttt{SumSquares} is smooth, while \texttt{L1Norm} and \texttt{Box} are nonsmooth but proxable.
Given a composite objective constructed via the \texttt{+} operator, \rlaopt{} partitions the terms:
\begin{itemize}
    \item All atoms with the smooth property form the smooth component $f$.
    \item All atoms without the smooth property form the nonsmooth components $g_1, \ldots, g_k$.
\end{itemize}
This partition is performed automatically and requires no input from the user.

\subsection{Splitting for Proximal Gradient Methods}
\label{subsec:rlaopt-prox-grad-split}
For proximal gradient and accelerated proximal gradient methods, \rlaopt{} requires that each nonsmooth atom $g_i$ acts directly on a variable (rather than on an affine expression of variables).
In this case, the proximal operator of $g_i$ can be applied directly to the variable at each iteration, yielding the standard proximal gradient update
\begin{equation}
    \label{eq:rlaopt-prox-grad-update}
    x^{k+1} = \prox_{\eta g}\!\left(x^k - \eta \nabla f(x^k)\right),
\end{equation}
where $\eta > 0$ is the step size and $g = \sum_{i=1}^k g_i$.

\rlaopt{} validates two conditions for proximal gradient splitting:
\begin{enumerate}
    \item Each nonsmooth atom must be proxable, meaning it takes a raw variable as input.
    \item The nonsmooth atoms must operate on \emph{disjoint} sets of variables, so that their proximal operators can be applied independently.
\end{enumerate}
If either condition is violated (for example, if a nonsmooth atom receives an affine expression as input), \rlaopt{} raises an error indicating that the problem structure is incompatible with proximal gradient methods and suggesting the use of ADMM instead.

\subsection{Splitting for ADMM}
\label{subsec:rlaopt-admm-split}
ADMM handles a broader class of problems than proximal gradient, as it allows nonsmooth atoms to receive affine expressions of the variables as input.
When a nonsmooth atom $g_i$ acts on an affine expression $A_i x - b_i$ rather than a raw variable, \rlaopt{} introduces an auxiliary variable $z_i$ and rewrites the problem in the consensus form
\begin{equation}
    \label{eq:rlaopt-admm-decomposition}
    \begin{array}{lll}
        \underset{x,\, z_1, \ldots, z_k}{\text{minimize}} & f(x) + \sum_{i=1}^{k} g_i(z_i) &                   \\
        \text{subject to}                                 & A_i x - z_i = b_i,             & i = 1, \ldots, k.
    \end{array}
\end{equation}
Each atom provides a \texttt{decompose} method that performs this transformation.
Given a nonsmooth atom $g_i$ with input expression $A_i x - b_i$, the decomposition produces:
\begin{enumerate}
    \item A new auxiliary variable $z_i$ whose shape matches the output dimension of $A_i x - b_i$.
    \item A new atom $g_i(z_i)$ that is proxable (since it now acts on a raw variable).
    \item The linear operator $A_i$ and offset $b_i$, extracted from the affine expression.
\end{enumerate}
The linear operator $A_i$ is represented implicitly as a \texttt{LinearOperator} object that computes matrix-vector products $v \mapsto A_i v$ and adjoint products $v \mapsto A_i^T v$ without forming $A_i$ explicitly.
This is important for scalability, as the linear operators arising from data matrices in machine learning can be very large.

The ADMM algorithm then alternates between approximately solving the $x$-subproblem (a smooth optimization problem involving $f$ and the quadratic penalty terms), applying the proximal operators of $g_i$ to update each $z_i$, and updating the dual variables.
The $x$-subproblem requires solving a linear system at each iteration, which \rlaopt{} accelerates using \nys{}-based preconditioning \citep{frangella2023randomized, zhao2022nysadmm}.

\subsection{Solver Selection}
Given a composite objective, the choice of solver depends on the problem structure.
\cref{tab:rlaopt-solver-selection} summarizes the conditions under which each solver family is applicable.

\begin{table}[t]
    \centering
    \small
    \begin{tabular}{lp{9cm}}
        \toprule
        \textbf{Solver}   & \textbf{Applicable when}                                                                                                         \\
        \midrule
        \nys{}PCG         & Problem is a pd linear system $Ax = b$.                                                                                          \\
        \midrule
        Proximal gradient & All nonsmooth atoms act on raw variables and operate on disjoint variable sets.                                                  \\
        \midrule
        NysADMM           & Nonsmooth atoms may act on affine expressions of variables. Automatically introduces auxiliary variables and linear constraints. \\
        \midrule
        SAPPHIRE          & Smooth component $f$ has a finite-sum structure $f(x) = \frac{1}{N}\sum_j f_j(x)$. All nonsmooth atoms act on raw variables.     \\
        \bottomrule
    \end{tabular}
    \caption{Conditions for solver applicability in \rlaopt{}.}
    \label{tab:rlaopt-solver-selection}
\end{table}

In practice, when the user specifies a solver, \rlaopt{} validates that the problem structure is compatible and raises an informative error if it is not.
This design gives users explicit control over the solver while preventing misuse through automatic structural checks.

\subsection{Example: Splitting in Action}
We illustrate the splitting process with a concrete example.
Consider the $\ell_1$-regularized least-squares problem where an affine transformation of the decision variable is penalized:
\begin{equation}
    \label{eq:rlaopt-splitting-example}
    \underset{w}{\text{minimize}} \quad \frac{1}{2N} \| X w - y \|_2^2 + \lambda \| C w - d \|_1,
\end{equation}
where $C \in \R^{m \times n}$ and $d \in \R^m$.
The user specifies this problem as:
\begin{CodeBlock}
\PYG{n}{w} \PYG{o}{=} \PYG{n}{Variable}\PYG{p}{(}\PYG{p}{(}\PYG{n}{n}\PYG{p}{,}\PYG{p}{)}\PYG{p}{)}
\PYG{n}{obj} \PYG{o}{=} \PYG{n}{SumSquares}\PYG{p}{(}\PYG{n}{X} \PYG{o}{@} \PYG{n}{w} \PYG{o}{\PYGZhy{}} \PYG{n}{y}\PYG{p}{)} \PYG{o}{*} \PYG{p}{(}\PYG{l+m+mf}{0.5} \PYG{o}{/} \PYG{n}{N}\PYG{p}{)} \PYG{o}{+} \PYG{n}{L1Norm}\PYG{p}{(}\PYG{n}{C} \PYG{o}{@} \PYG{n}{w} \PYG{o}{\PYGZhy{}} \PYG{n}{d}\PYG{p}{,} \PYG{n}{lam}\PYG{p}{)}
\end{CodeBlock}

For a proximal gradient solver, this problem is \emph{not} directly compatible, because the \texttt{L1Norm} atom receives the affine expression \texttt{C @ w - d} rather than a raw variable.
For ADMM, \rlaopt{} automatically detects this structure and decomposes the problem.
The \texttt{L1Norm} atom's \texttt{decompose} method introduces an auxiliary variable $z \in \R^m$ and rewrites the problem as
\begin{equation*}
    \begin{array}{ll}
        \underset{w,\, z}{\text{minimize}} & \frac{1}{2N} \| X w - y \|_2^2 + \lambda \| z \|_1 \\
        \text{subject to}                  & C w - z = d.
    \end{array}
\end{equation*}
The linear operator $C$ and offset $d$ are extracted from the affine expression, and the proximal operator of $\lambda \| \cdot \|_1$ (soft-thresholding) is applied to $z$ at each ADMM iteration.
The $w$-subproblem involves minimizing $\frac{1}{2N}\|Xw - y\|_2^2 + \frac{\rho}{2}\|Cw - z^k - d + u^k\|_2^2$, which amounts to solving a pd linear system that \rlaopt{} accelerates with \nys{} preconditioning.

\section{Experiments}
\label{sec:rlaopt-experiments}
We perform an extensive evaluation of \rlaopt{}'s solvers on large-scale problems, comparing these solvers with state-of-the-art alternatives.
We find that \rlaopt{}'s solvers are particularly effective on large, dense, ill-conditioned problems, where randomized preconditioning and GPU acceleration provide significant speedups.
On synthetic ridge regression, the benefit of \nys{}PCG depends on the conditioning of the linear system, reaching a $7 \times$ speedup over CG on the largest problem with fast spectral decay and weak regularization, while obtaining more favorable scaling compared to LSQR \citep{paige1982lsqr} and LSMR \citep{fong2011lsmr}.
On bounded multinomial logistic regression, SAPPHIRE does not outperform JAXopt's accelerated proximal gradient (APG) \citep{beck2009fast} and L-BFGS-B \citep{byrd1995limited} solvers when run on GPU, but could still be valuable for certain large-scale applications.
On bounded elastic net, NysADMM solves several large problems with dense data, while the competing conic solvers SCS \citep{odonoghue2016conic} and Clarabel \citep{chen2024clarabel,chen2025cuclarabel} either run out of memory or reach the time limit.
However, conic solvers are much faster than NysADMM when the data is sparse.
We conclude with a demonstration of the differentiable optimization capabilities of \rlaopt{}.
Although the solvers in \rlaopt{} do not universally outperform state-of-the-art solvers, we emphasize that the suite of solvers in \rlaopt{} can solve a much wider range of problems than any one of the solvers that we compare against.
Code for the experiments is available at \href{https://github.com/pratikrathore8/rlaopt-experiments}{https://github.com/pratikrathore8/rlaopt-experiments}.

Our experiments require solutions to satisfy a set of common accuracy checks, which may be more stringent than necessary for applications requiring only low-to-moderate precision.
These requirements may favor interior-point methods such as Clarabel and cuClarabel over first-order methods such as SCS and the methods in \rlaopt{}.
The relative performance of these solvers likely differs at looser accuracy levels.

\paragraph{Experimental setup.}
We use \rlaopt{} 0.1.0 by installing it from PyPI.
We compare solutions using common accuracy checks, independent of each solver's stopping criterion.
Ridge regression requires a relative residual less than $10^{-6}$; bounded multinomial regression and bounded elastic net require stationarity less than $10^{-4}$ and feasibility violation less than $10^{-6}$.
We use float64 arithmetic, 64 CPU cores and 128 GiB host memory per task, with one 141 GB NVIDIA H200 for GPU tasks.
The time limits are 900 seconds for ridge regression and 3600 seconds for bounded elastic net and multinomial logistic regression.
\cref{sec:experimental-protocol} gives additional details.

\subsection{Large-Scale Ridge Regression with \nys{}PCG}
We solve
\begin{equation}
 \min_{w\in\R^p}\ \frac12\|Xw-y\|_2^2+\frac{\lambda}{2}\|w\|_2^2,
 \qquad (X^T X+\lambda I)w=X^T y.
 \label{eq:experiment-ridge}
\end{equation}
To control the conditioning of the objective in \eqref{eq:experiment-ridge}, we construct
\begin{equation}
 X=U\Sigma_\alpha V^T,\qquad
 \Sigma_\alpha=\operatorname{diag}(i^{-\alpha/2})_{i=1}^r,\qquad
 y=Ug/\|g\|_2,\quad g\sim\mathcal N(0,I_r),
 \label{eq:experiment-ridge-data}
\end{equation}
where $r=\min(n,p)$, $U\in\R^{n\times r}$ and $V\in\R^{p\times r}$ have orthonormal columns.
We generate these columns from structured orthogonal matrices using the SORF construction of \citet[Eq.~(5)]{yu2016orthogonal}.
All problems have $n\geq p$, so the matrix $X^T X + \lambda I$ has eigenvalues $i^{-\alpha}+\lambda$ and condition number $(1+\lambda)/(p^{-\alpha}+\lambda)$.
Thus increasing $\alpha$ increases the condition number, while increasing $\lambda$ decreases the condition number.

We use $\alpha\in\{0.5,1,2\}$, $\lambda\in\{10^{-2},10^{-4},10^{-6}\}$, and three random seeds.
The dimension sweeps for $X$ comprise square matrices with $n=p\in\{2^{10},2^{12},2^{14},2^{16}\}$; fixed $p=2^{14}$ with $n\in\{2^{14},2^{15},2^{16}\}$; and fixed $n=2^{16}$ with $p\in\{2^{10},2^{12},2^{14},2^{16}\}$.
These sweeps cover eight distinct shapes for $X$ in total.
We compare rank-128 \nys{}PCG, CG \citep{hestenes1952methods}, QR, and LSQR on CPU, and replace LSQR with cuML's LSMR solver \citep{raschka2020machine} on GPU.
CG and \nys{}PCG apply the normal matrix through products with $X$ and $X^T$ without forming $X^T X$.
Preconditioner construction for \nys{}PCG is included in the measured solve time.

\begin{figure}[tbp]
 \centering\includegraphics[width=\linewidth]{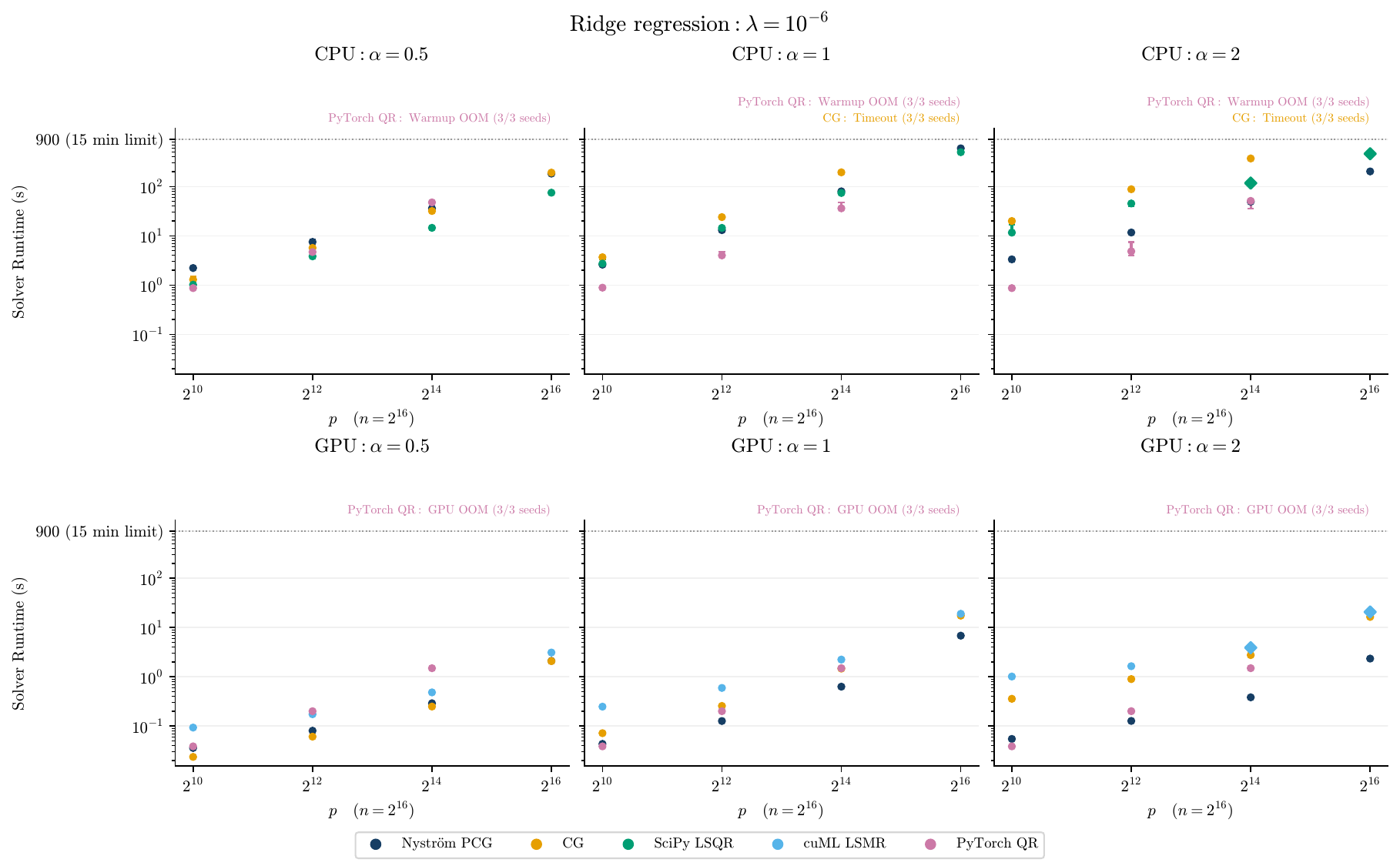}
 \caption{Ridge regression with $n=2^{16}$ and $\lambda=10^{-6}$, varying $p$ and spectral decay $\alpha$. Times include solver setup and, for \nys{}PCG, preconditioner construction. Each point on the plot indicates median solve time over three seeds; the error bars provide the min-max range of solve times. As the problems become more ill-conditioned (larger $\alpha$, smaller $\lambda$), \nys{}PCG scales better than the competing methods, especially on GPU.
 }
 \label{fig:ridge-scaling}

\end{figure}

\begin{figure}[tbp]
 \centering\includegraphics[width=\linewidth]{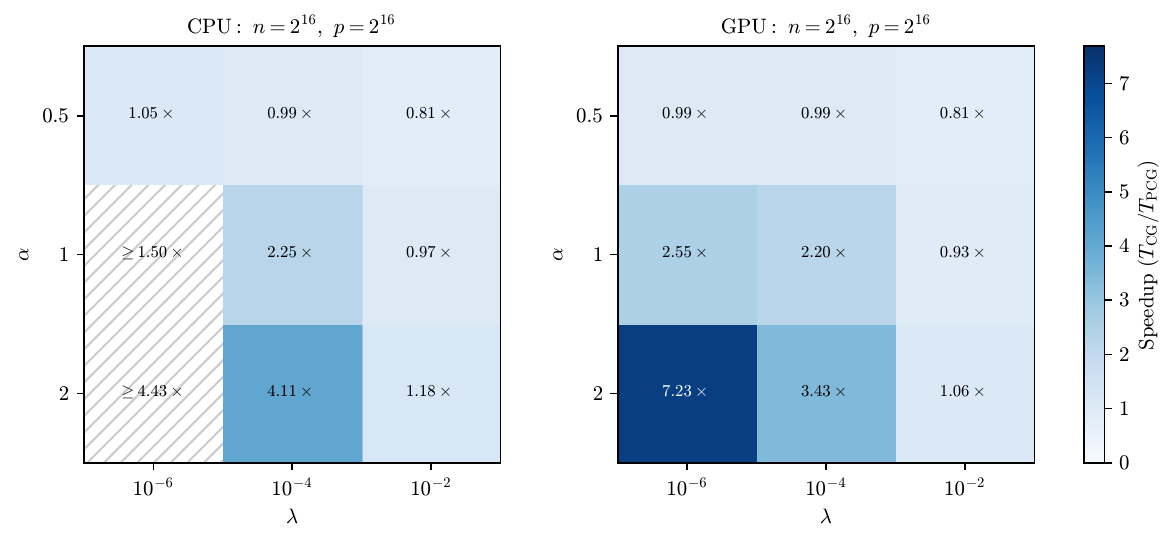}
 \caption{CG time divided by \nys{}PCG time at $n=p=2^{16}$, including preconditioner construction. Ratios above one favor preconditioning. Inequalities denote lower bounds when CG times out.}
 \label{fig:ridge-preconditioning}
\end{figure}

\cref{fig:ridge-scaling} shows the fixed-$n$ sweep at $\lambda=10^{-6}$.
\nys{}PCG on GPU is faster than on CPU by $57\times$--$128\times$.
The direct QR baseline encounters memory limits as the problem grows, and fails completely at $n=p=2^{16}$.
LSQR and LSMR outperform \nys{}PCG when the spectrum decays slowly or the regularization is strong, but \nys{}PCG is superior for ill-conditioned problems.
\cref{fig:ridge-preconditioning} isolates the effect of preconditioning at $n=p=2^{16}$.
For $\alpha=2$ and $\lambda=10^{-6}$, \nys{}PCG is $7.23\times$ faster than CG on GPU; at $\alpha=0.5$ and $\lambda=10^{-2}$, this ratio is merely $0.81$.
This is intuitive: the benefits of randomized preconditioning are greatest when the spectrum decays quickly (large $\alpha$) and the regularization is weak (small $\lambda$).
\cref{sec:additional-experiments} includes the remaining dimension and regularization sweeps.


\subsection{Bounded Multinomial Logistic Regression with SAPPHIRE}
We solve multinomial logistic regression:
\begin{equation}
 \min_{W\in\R^{p\times K}}\ -\frac1n\sum_{i=1}^n
 \log\bigl(\operatorname{softmax}(XW)_{i,c_i}\bigr)
 \quad\text{subject to}\quad -1\leq W\leq1.
 \label{eq:experiment-multinomial}
\end{equation}
We compare SAPPHIRE with accelerated projected gradient (APG) and L-BFGS-B from JAXopt.
The datasets are CIFAR-10, SVHN, Fashion-MNIST, News20, and RCV1; \cref{tab:experiment-datasets} lists their dimensions and preprocessing.
The main comparison disables JIT compilation in JAXopt for fairness, since \rlaopt{} is not currently JIT-enabled (\cref{fig:multinomial-jit} compares with JIT-enabled JAXopt).
Both configurations use the same common accuracy checks.

\begin{figure}[tbp]
 \centering\includegraphics[width=\linewidth]{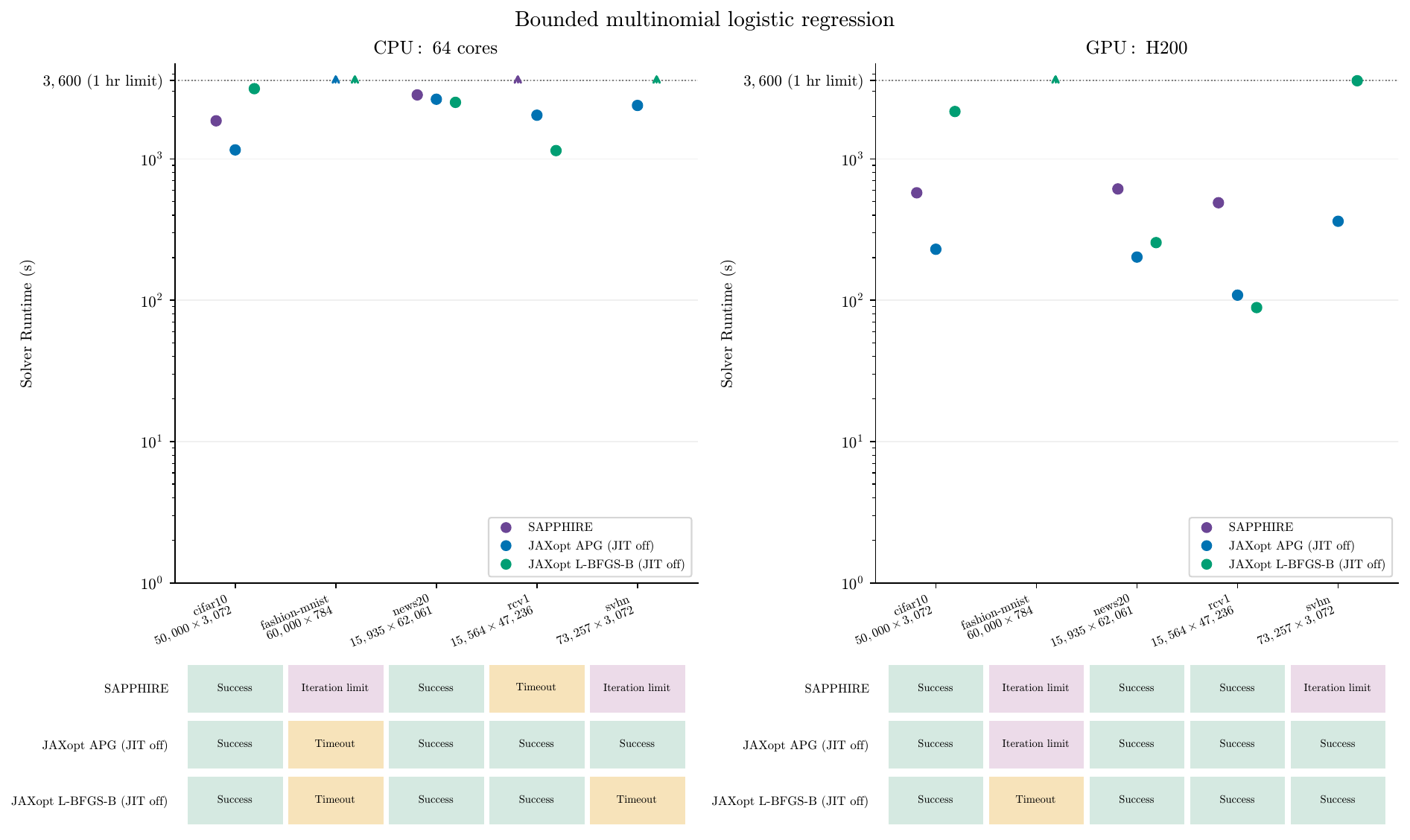}
 \caption{Bounded multinomial regression with JAXopt JIT compilation disabled. Each successful marker represents a solve satisfying stationarity $\leq10^{-4}$ and feasibility violation $\leq10^{-6}$. The JAXopt methods outperform SAPPHIRE in most instances. The JIT-enabled comparison appears in \cref{fig:multinomial-jit}.}
 \label{fig:bounded-multinomial}
\end{figure}

\cref{fig:bounded-multinomial} shows that GPU SAPPHIRE succeeds on CIFAR-10, News20, and RCV1, taking approximately 576, 613, and 490 seconds, respectively.
CPU SAPPHIRE succeeds on CIFAR-10 and News20 but times out on RCV1.
On SVHN and Fashion-MNIST, SAPPHIRE reaches the iteration limit.
APG is faster than SAPPHIRE on all three GPU instances on which SAPPHIRE succeeds.
JIT compilation further improves the JAXopt results: GPU L-BFGS-B succeeds on all five datasets, and both JIT-enabled baselines are faster than SAPPHIRE on its three successful instances.
Thus this experiment demonstrates GPU acceleration of SAPPHIRE, while also identifying a performance gap relative to these deterministic baselines.
Although SAPPHIRE is outperformed by JAXopt, it uses minibatch gradients for its updates and evaluates full gradients periodically to check termination.
For large, high-dimensional datasets where computing a full gradient at every iteration is too expensive, SAPPHIRE could be a more practical choice.

\subsection{Bounded Elastic Net with NysADMM}
We solve
\begin{equation}
 \min_{w\in\R^p,b\in\R}\ \frac{1}{2n}\|Xw+b\mathbf1-y\|_2^2
 +\lambda_1\|w\|_1+\frac{\lambda_2}{2}\|w\|_2^2
 \quad\text{subject to}\quad 0\leq w\leq1,
 \label{eq:experiment-bounded-en}
\end{equation}
with an unregularized intercept and $\lambda_1=\lambda_2=0.1\lambda_{\max}$, where
$\lambda_{\max}=\|X^T(y-\bar y\mathbf1)\|_\infty/n$.
We compare NysADMM \citep{zhao2022nysadmm} with SCS \citep{odonoghue2016conic} on both CPU and GPU (using both indirect and direct linear system solvers in the backend), Clarabel on CPU \citep{chen2024clarabel}, and cuClarabel on GPU \citep{chen2025cuclarabel}.
The datasets comprise three dense random-feature problems, acsincome-rf, yearpredictionmsd-rf, and yolanda-rf, and two sparse problems, e2006 and realsim.

\begin{figure}[tbp]
 \centering\includegraphics[width=\linewidth]{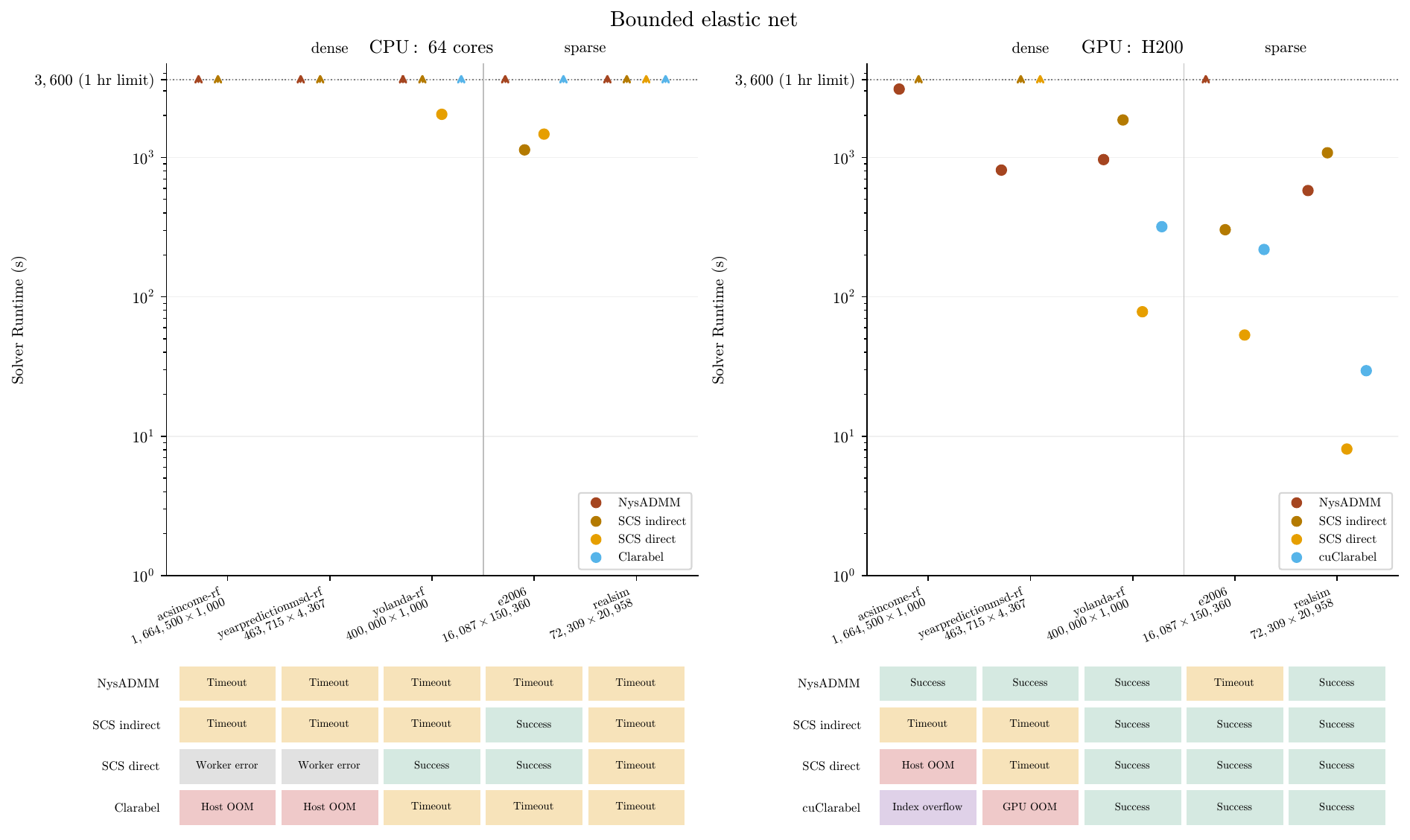}
 \caption{Bounded elastic net. Successful runs satisfy the common stationarity and feasibility checks. Timeout means that the displayed attempt exhausted its 3600 second budget. Host/GPU memory failures, index overflow, and worker errors are distinguished. The two CPU direct SCS worker errors are probably memory-related, but we cannot confirm this definitively. NysADMM is the only solver that succeeds on two of the large, dense random-feature problems, while SCS and Clarabel are superior on the sparse problems.}
 \label{fig:bounded-elastic-net}
\end{figure}

\cref{fig:bounded-elastic-net} shows that GPU NysADMM solves acsincome-rf and yearpredictionmsd-rf in approximately 3093 and 811 seconds, respectively, while no competing solver succeeds.
This advantage does not extend to every dataset.
On GPU, SCS direct solves yolanda-rf in 78 seconds and realsim in 8 seconds, compared with 965 and 579 seconds for NysADMM.
Moreover, SCS direct on GPU solves e2006 in 53 seconds, while NysADMM times out.
The results show that NysADMM is most effective on large, dense problems, while conic solvers are superior on sparse problems, even when they are high-dimensional.

\subsection{GPU Speedups}
\begin{figure}[tbp]
 \centering\includegraphics[width=\linewidth]{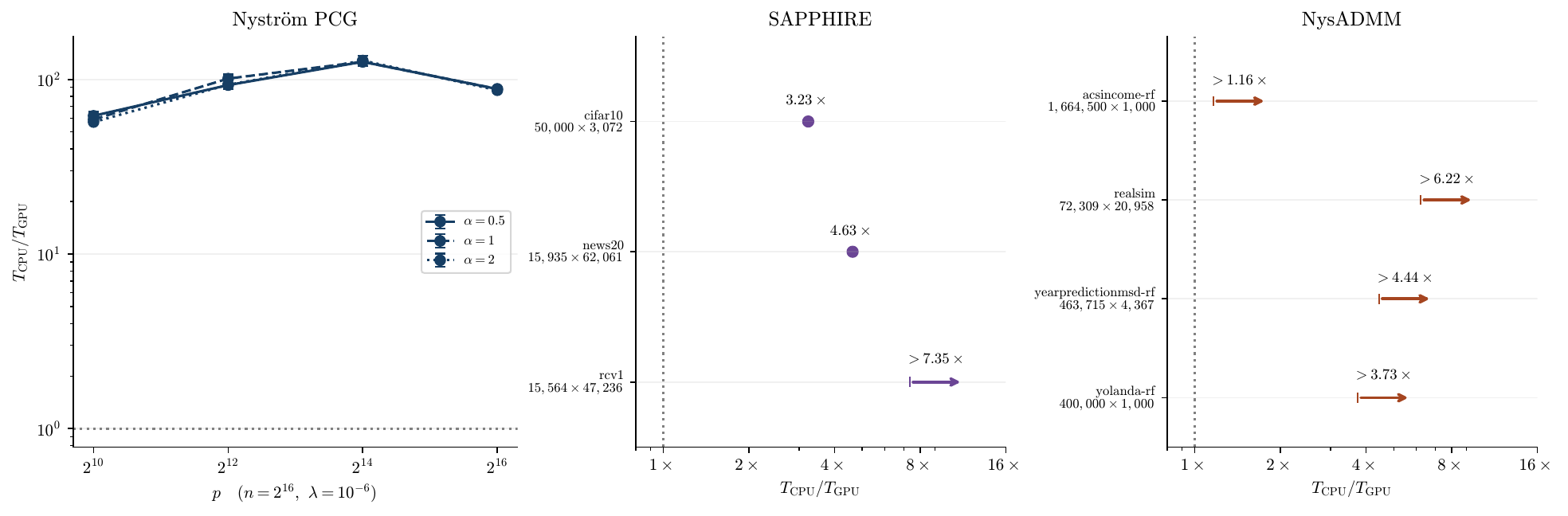}
 \caption{GPU vs. CPU speedups for \rlaopt{} solvers. Inequalities denote lower bounds obtained when the solver times out on CPU but succeeds on GPU. Problems where the solver fails on both CPU and GPU are not included.}
 \label{fig:rlaopt-gpu-speedups}
\end{figure}

\cref{fig:rlaopt-gpu-speedups} compares CPU and GPU times for the same solver and problem for the hardware systems described in \cref{sec:experimental-protocol}.
\nys{}PCG is $57$--$128\times$ faster on GPU than CPU for the synthetic ridge experiments, 
SAPPHIRE is $3.23\times$ faster on cifar10 and $4.63\times$ faster on news20, and its rcv1 speedup is $> 7.35\times$ because the CPU run times out.
The NysADMM speedups are lower bounds: CPU NysADMM times out on all five datasets, while its four successful GPU solves imply speedups exceeding $1.16$--$6.22\times$.

\subsection{Differentiating Through the Solver}
Finally, we demonstrate \rlaopt{}'s ability to differentiate through the optimization solver.
We consider the task of tuning the regularization parameter $\mu$ in a lasso problem to minimize the prediction error on a held-out validation set, as formulated in \eqref{eq:rlaopt-lasso-tuning}.

\begin{figure}[t]
    \centering
    \includegraphics[width=0.7\linewidth]{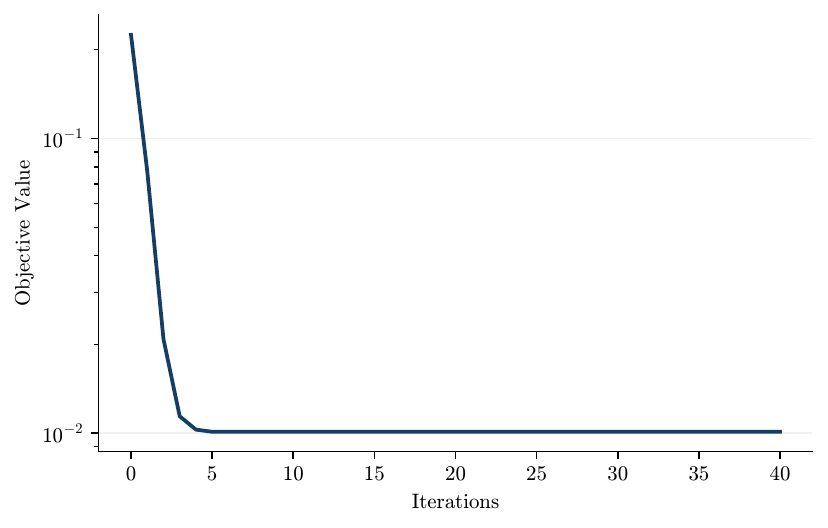}
    \caption[Differentiating through \rlaopt{}'s proximal gradient solver enables gradient-based tuning of the lasso regularization parameter $\mu$.]{Differentiating through \rlaopt{}'s proximal gradient solver enables gradient-based tuning of the lasso regularization parameter $\mu$. The validation loss decreases as $\mu$ is optimized.}
    \label{fig:rlaopt-diff-solver}
\end{figure}

\cref{fig:rlaopt-diff-solver} shows the convergence of the outer gradient descent loop that optimizes $\mu$.
At each outer iteration, \rlaopt{} solves the inner lasso problem using proximal gradient and then computes the gradient of the validation loss with respect to $\mu$ via PyTorch's automatic differentiation.
The validation loss decreases steadily, demonstrating that \rlaopt{} correctly propagates gradients through the solver and enables effective hyperparameter tuning.

\section{Conclusion}
\label{sec:rlaopt-conclusion}
We have developed \rlaopt{}, an open-source, PyTorch-based software package for large-scale optimization using randomized linear algebra.
\rlaopt{} addresses a significant gap between the algorithmic foundations of RandNLA and the software available to practitioners, providing GPU-enabled implementations of \nys{}PCG, NysADMM, and SAPPHIRE within a unified framework.
We have introduced a flexible modeling language inspired by CVXPY that allows users to specify optimization problems using natural mathematical syntax, and described how \rlaopt{} automatically detects problem structure to perform the appropriate decomposition for each solver.
Our experiments show substantial GPU speedups and demonstrate that the benefit of randomized preconditioning depends on spectral decay and regularization. 
GPU NysADMM solves two dense bounded elastic-net instances on which no competing conic solver succeeds in our experiments, while conic solvers are faster on sparse instances. 
SAPPHIRE benefits from GPU execution, although JAXopt baselines are faster on the multinomial instances that SAPPHIRE solves. 
We also demonstrate differentiation through the solver, enabling gradient-based hyperparameter tuning.

There are three promising directions for future work.
First, we can expand the set of supported atoms and solvers, including NysNewton-CG \citep{rathore2024challenges}, which combines \nys{}PCG with Newton's method, and ASkotch \citep{rathore2026askotch}, which combines RandNLA with sketch-and-project solvers \citep{gower2015randomized} for linear systems.
Second, we can extend \rlaopt{} to support more scientific computing applications from RandNLA, including stochastic trace estimation \citep{hutchinson1990stochastic, meyer2021hutchplusplus} and spectral density estimation \citep{lin2016approximating, ubaru2017fast, yao2020pyhessian}.
Finally, we can improve performance on sparse matrices and explore a JAX-based backend with JIT compilation.

\rlaopt{} is available at \href{https://github.com/udellgroup/rlaopt}{https://github.com/udellgroup/rlaopt}, with documentation at \href{https://rlaopt.readthedocs.io}{https://rlaopt.readthedocs.io} and version 0.1.0 on \href{https://pypi.org/project/rlaopt/0.1.0/}{PyPI}.

\section*{Acknowledgments}
PR, ZF, and MU gratefully acknowledge support from
the Office of Naval Research under award N000142412306, % RandNLA for optimization
Air Force Office of Scientific Research under award FA9550-26-1-0012, % OSGM
the Alfred P. Sloan Foundation,
the Stanford Institute for Human-Centered Artificial Intelligence (HAI),
and from IBM Research as a founding member of Stanford Institute for Human-centered Artificial Intelligence.
We would also like to acknowledge the SC cluster hosted by Stanford Computer Science, which provided the resources to run the experiments in this paper.

\bibliographystyle{ACM-Reference-Format}
\bibliography{references}

\clearpage
\appendix

\section{Solver Configuration and Termination Criteria}
\label{app:rlaopt-solver-configs}
This appendix provides a complete reference for the configuration parameters and termination criteria of each solver in \rlaopt{} 0.1.0.
All configuration classes use sensible defaults, allowing users to get started with minimal tuning.

\subsection{\nys{}PCG Configuration}
\label{app:rlaopt-pcg-config}

\paragraph{Solver configuration (\texttt{PCGConfig}).}
\cref{tab:rlaopt-pcg-config} lists the parameters of \texttt{PCGConfig}.
The default \texttt{IdentityConfig} gives ordinary CG; using \texttt{NystromConfig} gives \nys{}PCG.

\begin{table}[htbp]
    \centering
    \small
    \begin{tabular}{llll}
        \toprule
        \textbf{Parameter}              & \textbf{Type}                 & \textbf{Default}          & \textbf{Description}     \\
        \midrule
        \texttt{preconditioner\_config} & \texttt{PreconditionerConfig} & \texttt{IdentityConfig()} & Preconditioner strategy. \\
        \bottomrule
    \end{tabular}
    \caption{\texttt{PCGConfig} parameters.}
    \label{tab:rlaopt-pcg-config}
\end{table}

\paragraph{Stopping criteria (\texttt{PCGStoppingCriteria}).}
\cref{tab:rlaopt-pcg-stopping} lists the termination parameters for \nys{}PCG.
The solver terminates when the relative residual satisfies
\[
    \| r_k \|_2 \leq \texttt{tol} \cdot \| b \|_2,
\]
where $r_k = b - A x_k$ is the residual at iteration $k$, or when the iteration count reaches \texttt{max\_iters}.
For multiple right-hand sides, the relative residual condition must hold for each right-hand side.

\begin{table}[htbp]
    \centering
    \small
    \begin{tabular}{llll}
        \toprule
        \textbf{Parameter}  & \textbf{Type}  & \textbf{Default} & \textbf{Description}                \\
        \midrule
        \texttt{max\_iters} & \texttt{int}   & 1000             & Maximum number of iterations.       \\
        \texttt{tol}        & \texttt{float} & $10^{-6}$        & Relative tolerance for convergence. \\
        \bottomrule
    \end{tabular}
    \caption{\texttt{PCGStoppingCriteria} parameters.}
    \label{tab:rlaopt-pcg-stopping}
\end{table}

\subsection{NysADMM Configuration}
\label{app:rlaopt-admm-config}

\paragraph{Solver configuration (\texttt{ADMMConfig}).}
\cref{tab:rlaopt-admm-config} lists the parameters of \texttt{ADMMConfig}.
The NysADMM solver is accessed via the \texttt{ADMM} class in the API.
Its default preconditioner is \texttt{NystromConfig(rank\_init=50, base\_damping=0.0)}; the remaining parameters use the defaults in \cref{tab:rlaopt-nystrom-config}.

\begin{table}[htbp]
    \centering
    \small
    \setlength{\tabcolsep}{3pt}
    \begin{tabular}{@{}>{\raggedright\arraybackslash}p{0.29\linewidth}>{\raggedright\arraybackslash}p{0.20\linewidth}>{\raggedright\arraybackslash}p{0.16\linewidth}>{\raggedright\arraybackslash}p{0.28\linewidth}@{}}
        \toprule
        \textbf{Parameter}                    & \textbf{Type}                 & \textbf{Default}       & \textbf{Description}                                                           \\
        \midrule
        \texttt{rho}                          & \texttt{float}                & 1.0                    & Augmented Lagrangian penalty parameter.                                        \\
        \texttt{rho\_\allowbreak update\_\allowbreak factor}          & \texttt{float}                & 2.0                    & Multiplicative factor for updating $\rho$ during primal-dual balancing.        \\
        \texttt{rho\_\allowbreak update\_\allowbreak threshold}       & \texttt{float}                & 10.0                   & Threshold ratio of primal to dual residual that triggers an update to $\rho$.      \\
        \texttt{rho\_\allowbreak update\_\allowbreak freq}            & \texttt{int}                  & 25                     & Frequency (in iterations) for checking and updating $\rho$.                    \\
        \texttt{alpha}                        & \texttt{float}                & 1.6                    & Over-relaxation parameter ($0 < \alpha < 2$).                                  \\
        \texttt{sigma}                        & \texttt{float}                & $10^{-6}$              & Regularization for the inexact linear system solve.                            \\
        \texttt{gamma}                        & \texttt{float}                & 1.2                    & Exponent controlling the decay of the linear system solve tolerance ($> 1$).   \\
        \texttt{preconditioner\_\allowbreak config}       & \texttt{Preconditioner\-Config} & \texttt{NystromConfig} & Preconditioner for the linear system subproblem. Default: \nys{} with rank 50. \\
        \texttt{preconditioner\_\allowbreak update\_\allowbreak freq} & \texttt{int}                  & 20                     & Frequency (in iterations) for updating the preconditioner.                     \\
        \bottomrule
    \end{tabular}
    \caption{\texttt{ADMMConfig} parameters.}
    \label{tab:rlaopt-admm-config}
\end{table}

\paragraph{Stopping criteria (\texttt{ADMMStoppingCriteria}).}
\cref{tab:rlaopt-admm-stopping} lists the termination parameters for NysADMM.
The primal and dual residuals at iteration $k$ are
\[
    r_k^{\text{pri}} = Ax_k-z_k-b,
    \qquad r_k^{\text{dual}} = \nabla f(x_k)+\rho_k A^T u_k,
\]
where $u_k$ is the scaled dual variable and $\rho_k$ is the penalty parameter.
The solver terminates when both residuals fall below their respective tolerances,
\[
    \|r_k^{\text{pri}}\|_2 \leq \epsilon_k^{\text{pri}}
    \quad\text{and}\quad
    \|r_k^{\text{dual}}\|_2 \leq \epsilon_k^{\text{dual}},
\]
where
\[
    \begin{aligned}
    \epsilon_k^{\text{pri}} &= \sqrt{m}\,\epsilon_{\text{abs}} + \epsilon_{\text{rel}}\max\!\big(\|Ax_k\|_2,\,\|z_k\|_2,\,\|b\|_2\big),\\
    \epsilon_k^{\text{dual}} &= \sqrt{n}\,\epsilon_{\text{abs}} + \epsilon_{\text{rel}}\|\rho_k A^T u_k\|_2.
    \end{aligned}
\]
Here $m$ is the number of scalar constraints in the ADMM splitting and $n$ is the number of scalar decision variables.
These tolerances follow the standard ADMM convergence criterion \citep{boyd2011distributed}.
The solver also terminates when the iteration count reaches \texttt{max\_iters}.

\begin{table}[htbp]
    \centering
    \small
    \begin{tabular}{llll}
        \toprule
        \textbf{Parameter}  & \textbf{Type}  & \textbf{Default} & \textbf{Description}                              \\
        \midrule
        \texttt{max\_iters} & \texttt{int}   & 1000             & Maximum number of iterations.                     \\
        \texttt{eps\_abs}   & \texttt{float} & $10^{-4}$        & Absolute tolerance for primal and dual residuals. \\
        \texttt{eps\_rel}   & \texttt{float} & $10^{-4}$        & Relative tolerance for primal and dual residuals. \\
        \bottomrule
    \end{tabular}
    \caption{\texttt{ADMMStoppingCriteria} parameters.}
    \label{tab:rlaopt-admm-stopping}
\end{table}

\subsection{Proximal Gradient Configuration}
\label{app:rlaopt-prox-grad-config}

\paragraph{Solver configuration (\texttt{ProxGradConfig}).}
\cref{tab:rlaopt-prox-grad-config} lists the parameters of \texttt{ProxGradConfig}.
When the preconditioner is not the identity, both \texttt{use\_linesearch} and \texttt{use\_acceleration} must be \texttt{False}.
Line search and automatic stepsize updates cannot be enabled at the same time.
When the objective has a nonsmooth term and the preconditioner is non-identity, \texttt{subproblem\_iters} controls the number of accelerated proximal gradient iterations used to approximate the scaled proximal operator.
This parameter is ignored when the objective is smooth or the preconditioner is the identity.

\begin{table}[htbp]
    \centering
    \small
    \setlength{\tabcolsep}{3pt}
    \begin{tabular}{@{}>{\raggedright\arraybackslash}p{0.29\linewidth}>{\raggedright\arraybackslash}p{0.20\linewidth}>{\raggedright\arraybackslash}p{0.16\linewidth}>{\raggedright\arraybackslash}p{0.28\linewidth}@{}}
        \toprule
        \textbf{Parameter} & \textbf{Type} & \textbf{Default} & \textbf{Description} \\
        \midrule
        \texttt{eta} & \texttt{float} & 1.0 & Step size for the gradient update. \\
        \texttt{use\_\allowbreak acceleration} & \texttt{bool} & \texttt{False} & Whether to use Nesterov acceleration. \\
        \texttt{use\_\allowbreak linesearch} & \texttt{bool} & \texttt{True} & Whether to use backtracking line search. \\
        \texttt{precond\_\allowbreak config} & \texttt{Preconditioner\-Config} & \texttt{IdentityConfig()} & Preconditioner strategy. \\
        \texttt{subproblem\_\allowbreak iters} & \texttt{int} & 20 & Iterations for the scaled proximal solve. \\
        \texttt{auto\_\allowbreak update\_\allowbreak stepsize} & \texttt{bool} & \texttt{False} & Whether to estimate the step size from local curvature. \\
        \texttt{precond\_\allowbreak update\_\allowbreak freq} & \texttt{int} & 10 & Iterations between preconditioner and automatic stepsize updates. \\
        \bottomrule
    \end{tabular}
    \caption{\texttt{ProxGradConfig} parameters.}
    \label{tab:rlaopt-prox-grad-config}
\end{table}

\paragraph{Stopping criteria (\texttt{GradSolverStoppingCriteria}).}
\cref{tab:rlaopt-prox-grad-stopping} lists the termination parameters shared by proximal gradient and SAPPHIRE.
The proximal gradient solver computes the gradient mapping after every iteration and terminates when
\begin{equation}
    \label{eq:rlaopt-grad-stopping}
    \frac{1}{\eta_k}\|x_k-\prox_{\eta_k g}(x_k-\eta_k\nabla f(x_k))\|_2
    \leq \epsilon_{\text{abs}}+\epsilon_{\text{rel}}\|x_k\|_2,
\end{equation}
where $\eta_k$ is the current step size, $\epsilon_{\text{abs}}$ and $\epsilon_{\text{rel}}$ are the absolute and relative tolerances, and $\prox_{\eta_k g}$ denotes the proximal operator of $\eta_k g$.
The left-hand side of \eqref{eq:rlaopt-grad-stopping} vanishes at the optimum; for a smooth objective, the left-hand side reduces to the gradient norm.
The solver also terminates when the iteration count reaches \texttt{max\_iters}.

\begin{table}[htbp]
    \centering
    \small
    \setlength{\tabcolsep}{3pt}
    \begin{tabular}{@{}>{\raggedright\arraybackslash}p{0.29\linewidth}>{\raggedright\arraybackslash}p{0.20\linewidth}>{\raggedright\arraybackslash}p{0.16\linewidth}>{\raggedright\arraybackslash}p{0.28\linewidth}@{}}
        \toprule
        \textbf{Parameter} & \textbf{Type} & \textbf{Default} & \textbf{Description} \\
        \midrule
        \texttt{max\_\allowbreak iters} & \texttt{int} & 1000 & Maximum number of iterations (minibatch updates for SAPPHIRE). \\
        \texttt{eps\_\allowbreak abs} & \texttt{float} & $10^{-4}$ & Absolute tolerance for the gradient mapping norm. \\
        \texttt{eps\_\allowbreak rel} & \texttt{float} & $10^{-4}$ & Relative tolerance for the gradient mapping norm. \\
        \bottomrule
    \end{tabular}
    \caption{\texttt{GradSolverStoppingCriteria} parameters for proximal gradient and SAPPHIRE.}
    \label{tab:rlaopt-prox-grad-stopping}
\end{table}

\subsection{SAPPHIRE Configuration}
\label{app:rlaopt-sapphire-config}

\paragraph{Solver configuration (\texttt{SapphireConfig}).}
\cref{tab:rlaopt-sapphire-config} lists the parameters of \texttt{SapphireConfig}.
The SAPPHIRE solver is accessed via the \texttt{Sapphire} class in the API.
The base method can be \texttt{"saga"}, \texttt{"svrg"}, or \texttt{"sgd"}; the default is \texttt{"saga"}.
The default preconditioner uses \texttt{NystromConfig} with \texttt{rank\_init=10}, \texttt{error\_tolerance=0.1}, \texttt{base\_damping=0.001}, and \texttt{damping\_mode="adaptive"}.
The default maximum rank resolves to 10; other parameters use the defaults in \cref{tab:rlaopt-nystrom-config}.
Automatic stepsize updates are enabled by default; to use a fixed stepsize \texttt{eta}, set \texttt{auto\_update\_stepsize=False}.
The \texttt{subproblem\_iters} parameter has the same role as in proximal gradient.

\begin{table}[htbp]
    \centering
    \small
    \setlength{\tabcolsep}{3pt}
    \begin{tabular}{@{}>{\raggedright\arraybackslash}p{0.29\linewidth}>{\raggedright\arraybackslash}p{0.20\linewidth}>{\raggedright\arraybackslash}p{0.16\linewidth}>{\raggedright\arraybackslash}p{0.28\linewidth}@{}}
        \toprule
        \textbf{Parameter} & \textbf{Type} & \textbf{Default} & \textbf{Description} \\
        \midrule
        \texttt{base\_\allowbreak method} & \texttt{str} & \texttt{"saga"} & Base stochastic method: SAGA, SVRG, or SGD. \\
        \texttt{eta} & \texttt{float} & 0.1 & Fixed step size when automatic updates are disabled. \\
        \texttt{precond\_\allowbreak config} & \texttt{Preconditioner\-Config} & \texttt{NystromConfig} & \nys{} preconditioner with the defaults specified in the text. \\
        \texttt{subproblem\_\allowbreak iters} & \texttt{int} & 20 & Iterations for the scaled proximal solve. \\
        \texttt{auto\_\allowbreak update\_\allowbreak stepsize} & \texttt{bool} & \texttt{True} & Whether to estimate the step size from local curvature. \\
        \texttt{precond\_\allowbreak update\_\allowbreak freq} & \texttt{int} & 2 & Epochs between preconditioner and automatic stepsize updates. \\
        \texttt{snapshot\_\allowbreak update\_\allowbreak freq} & \texttt{int} & 1 & Epochs between snapshot updates (SVRG only). \\
        \texttt{check\_\allowbreak termination\_\allowbreak freq} & \texttt{int} & 1 & Epochs between gradient-mapping evaluations. \\
        \bottomrule
    \end{tabular}
    \caption{\texttt{SapphireConfig} parameters.}
    \label{tab:rlaopt-sapphire-config}
\end{table}

The minibatch size $B$ is specified through the model's \texttt{DataLoader}, rather than \texttt{SapphireConfig}.
For the update frequencies, the implementation converts one epoch to $\lfloor N/B\rfloor$ minibatch updates, where $N$ is the number of training samples.
The preconditioner is initialized on the first update and refreshed at the configured interval; snapshot updates apply only to SVRG.

\paragraph{Stopping criteria (\texttt{GradSolverStoppingCriteria}).}
SAPPHIRE uses the parameters in \cref{tab:rlaopt-prox-grad-stopping} and the gradient-mapping condition in \eqref{eq:rlaopt-grad-stopping}.
The solver evaluates this mapping using a full gradient after the first minibatch update and every \texttt{check\_termination\_freq} epochs thereafter.
Between these evaluations, it retains the most recently computed error for assessing convergence.
The solver also terminates after \texttt{max\_iters} minibatch updates.

\subsection{\nys{} Preconditioner Configuration}
\label{app:rlaopt-nystrom-config}

\paragraph{Preconditioner configuration (\texttt{NystromConfig}).}
\cref{tab:rlaopt-nystrom-config} lists the parameters of \texttt{NystromConfig}, which controls the randomized \nys{} preconditioner used by \nys{}PCG, NysADMM, SAPPHIRE, and optionally proximal gradient.

\begin{table}[htbp]
    \centering
    \small
    \setlength{\tabcolsep}{3pt}
    \begin{tabular}{@{}>{\raggedright\arraybackslash}p{0.29\linewidth}>{\raggedright\arraybackslash}p{0.20\linewidth}>{\raggedright\arraybackslash}p{0.16\linewidth}>{\raggedright\arraybackslash}p{0.28\linewidth}@{}}
        \toprule
        \textbf{Parameter}         & \textbf{Type}                 & \textbf{Default}    & \textbf{Description}                                                                                                           \\
        \midrule
        \texttt{rank\_\allowbreak init}        & \texttt{int}                  & (required)          & Initial rank of the \nys{} approximation.                                                                                      \\
        \texttt{rank\_\allowbreak max}         & \texttt{int} or \texttt{None} & \texttt{None}       & Maximum allowable rank. Defaults to \texttt{rank\_\allowbreak init} if not specified.                                                      \\
        \texttt{num\_\allowbreak power\_\allowbreak iters} & \texttt{int}                  & 10                  & Number of power iterations for error estimation in rank adaptation.                                                            \\
        \texttt{error\_\allowbreak tolerance}  & \texttt{float}                & $10^{-2}$           & Error tolerance for rank adaptation.                                                                                           \\
        \texttt{base\_\allowbreak damping}     & \texttt{float}                & (required)          & Base damping for the regularized low-rank approximation; required and nonnegative.                                                \\
        \texttt{damping\_\allowbreak mode}     & \texttt{str}                  & \texttt{"adaptive"} & \texttt{"adaptive"}: adds the smallest retained approximate eigenvalue to \texttt{base\_\allowbreak damping}. \texttt{"non\_\allowbreak adaptive"}: uses \texttt{base\_\allowbreak damping} only. \\
        \bottomrule
    \end{tabular}
    \caption{\texttt{NystromConfig} parameters.}
    \label{tab:rlaopt-nystrom-config}
\end{table}

\section{Experimental Details}
\label{sec:experimental-protocol}
This appendix provides details and settings for the experiments in \cref{sec:rlaopt-experiments}.

\subsection{Datasets and Preprocessing}
\label{sec:experiment-preprocessing}
\cref{tab:experiment-datasets} summarizes the training data used in the bounded problems.
We use LIBSVM for yearpredictionmsd, e2006, realsim, cifar10, svhn, news20, and rcv1, (\url{https://www.csie.ntu.edu.tw/~cjlin/libsvmtools/datasets/}), and OpenML (\url{https://www.openml.org/}) for acsincome (data ID 43141), yolanda (42705), and fashion-mnist (40996).
For fashion-mnist, we use the first 60000 examples of the 70000-example OpenML distribution as the training set.
Class labels are mapped to consecutive integers.
The experiments measure optimization on the training objective only.

\begin{table}[t]
\centering\small
\caption{Training dimensions and feature preprocessing. The column $d$ gives the source feature dimension and $p$ the preprocessed feature dimension. Standardization transforms the columns to have zero mean and unit variance; normalization divides each nonzero row by its Euclidean norm. $K$ is the number of classes for multinomial logistic regression.}
\label{tab:experiment-datasets}
\begin{tabular}{lrrrl}
\toprule
Dataset & $n$ & $d$ & $p$ & Feature preprocessing \\
\midrule
ACSIncome-rf & 1664500 & 11 & 1000 & Standardize; Gaussian features \\
Yolanda-rf & 400000 & 100 & 1000 & Standardize; Gaussian features \\
YearPredictionMSD-rf & 463715 & 90 & 4367 & Normalize; ReLU features \\
E2006-tfidf & 16087 & 150360 & 150360 & Normalize \\
Real-sim & 72309 & 20958 & 20958 & Normalize \\
\midrule
CIFAR-10 ($K=10$) & 50000 & 3072 & 3072 & Normalize \\
SVHN ($K=10$) & 73257 & 3072 & 3072 & Normalize \\
Fashion-MNIST ($K=10$) & 60000 & 784 & 784 & Standardize \\
News20 ($K=20$) & 15935 & 62061 & 62061 & Normalize \\
RCV1 ($K=51$) & 15564 & 47236 & 47236 & Normalize \\
\bottomrule
\end{tabular}
\end{table}

acsincome and yolanda's targets are centered and divided by their population standard deviation; other targets retain their source values.
In particular, realsim's source labels serve as regression targets in the bounded elastic-net experiment.
Sparse features remain sparse whenever the solver supports sparse matrices, while \rlaopt{} forces them to be dense.

\paragraph{Random features.}
We follow the random features implementation from \citet{frangella2024promise}.
For output dimension $m=p$, draw $G\in\R^{m\times d}$ with independent standard normal entries and set $W=G/\sqrt m$.
Gaussian features use
$\phi(x)=\sqrt{2/m}\cos(Wx/\sigma+\theta)$ with $\sigma=1$ and independent $\theta_j\sim\operatorname{Unif}[0,2\pi]$.
ReLU features use $\phi(x)=\max(Wx,0)$, coordinatewise.

\subsection{Common Accuracy Checks}
\label{sec:experiment-accuracy}
All accuracy checks are computed in float64 from the primal variables in the optimization problem: they do not use dual variables.
For ridge regression, the relative residual is
\begin{equation}
 \frac{\|(X^T X+\lambda I)w-X^T y\|_2}{\|X^T y\|_2}\leq10^{-6}.
 \label{eq:experiment-relative-residual}
\end{equation}
For the bounded problems, the accuracy checks measure stationarity with respect to the box and primal feasibility.
Let $(a)_+=\max(a,0)$ and $\delta=10^{-6}$.
For one coordinate $z$ with gradient $g$ and bounds $[\ell,u]$, define
\begin{equation}
 v_\delta(z,g;\ell,u)=
 \begin{cases}
 (-g)_+, & z\leq\ell+\delta,\\
 (g)_+, & z\geq u-\delta,\\
 |g|, & \text{otherwise}.
 \end{cases}
 \label{eq:experiment-box-stationarity}
\end{equation}
At a lower bound a nonnegative gradient is stationary, and at an upper bound a nonpositive gradient is stationary.
Feasibility is checked separately, so an infeasible coordinate cannot qualify solely because of this stationarity convention.

\paragraph{Bounded multinomial logistic regression.}
Let $P=\operatorname{softmax}(XW)$ rowwise, let $Y$ be the one-hot label matrix, and set $G=X^T(P-Y)/n$.
The stationarity and feasibility measures are
\begin{equation}
 s=\max_{j,k}v_\delta(W_{jk},G_{jk};-1,1),\qquad
 f=\max_{j,k}\{(-1-W_{jk})_+,(W_{jk}-1)_+\}.
\end{equation}

\paragraph{Bounded elastic net.}
Let $r=Xw+b\mathbf1-y$.
Because $w\geq0$, the $\ell_1$ penalty is linear on the feasible set, so we use
$g=X^T r/n+\lambda_2w+\lambda_1\mathbf1$.
The checks are
\begin{equation}
 s=\max\left\{\max_j v_\delta(w_j,g_j;0,1),
 \left|\frac{\mathbf1^T r}{n}\right|\right\},\qquad
 f=\max_j\{(-w_j)_+,(w_j-1)_+\}.
\end{equation}
Both bounded problems require $s\leq10^{-4}$ and $f\leq10^{-6}$.

\subsection{Calibration and Production Refinement}
\label{sec:experimental-refinement}
Tolerances are not immediately comparable between solvers because their stopping criteria and scaling conventions differ.
Calibration searches for the loosest native tolerance (for a single solver) whose solutions pass the common accuracy checks (\cref{sec:experiment-accuracy}) on every \textit{calibration instance} for a solver and backend.
The purpose of calibration is to reduce the number of runs on the large datasets used in the experiments, while the common accuracy checks remain the criterion for accepting a production result.
Calibration for ridge regression uses shapes $(2^{8},2^{8})$, $(2^{12},2^{12})$, and $(2^{14},2^{12})$, and all three values of $\alpha$ and $\lambda$ used in the experiments.
The calibration for bounded elastic net and multinomial logistic regression uses synthetic standardized Gaussian data with $(n,p)\in\{(2^{8},2^{5}),(2^{10},2^{6}),(2^{12},2^{8})\}$, five classes for multinomial logistic regression, and candidate tolerances $10^{-4},\ldots,10^{-10}$.
Elastic net calibration includes regularization fractions 0.1 and 0.01.
These runs use a 900-second limit and at most 10000 iterations.
\cref{tab:experiment-tolerances} records the solver tolerances obtained from the calibration procedure.

\begin{table}[t]
\centering
\caption{Solver tolerances obtained from the calibration procedure. Each value applies on both backends (CPU and/or GPU) where the method is available.}
\label{tab:experiment-tolerances}
\begin{tabular}{lll}
\toprule
Problem & Solver & Native tolerance \\
\midrule
Ridge & CG, \nys{}PCG & $10^{-6}$ \\
Ridge & SciPy LSQR, cuML LSMR & $10^{-9}$ \\
Ridge & QR & N/A \\
Bounded multinomial logistic& SAPPHIRE & $10^{-7}$ \\
Bounded multinomial logistic & JAXopt APG, L-BFGS-B & $10^{-6}$ \\
Bounded elastic net & NysADMM, all SCS backends & $10^{-7}$ \\
Bounded elastic net & Clarabel, cuClarabel & $10^{-10}$ \\
\bottomrule
\end{tabular}
\end{table}

\paragraph{Refinement rule.}
When running the experiments in the paper, we use the calibrated tolerances as the stopping criterion for each solver.
However, some solvers may terminate successfully according to their calibrated tolerance but fail the common accuracy checks.
For example, cuML LSMR may find a solution with its tolerance set to $10^{-9}$, but still fail to obtain a residual less than $10^{-6}$.
When this occurs, we rerun the solver with a stricter tolerance.
The runtimes given in the main paper are the runtimes of the first attempt that reaches its calibrated tolerance (perhaps after a rerun with a strict tolerance) and passes the common accuracy checks.
If no attempt qualifies, the plot displays the final completed attempt's failure outcome.

\paragraph{Ridge outcomes.}
Six CPU LSQR and seven GPU cuML LSMR runs completed but had relative residuals between $1.05\times10^{-6}$ and $3.09\times10^{-6}$.
All 13 passed after reducing their native tolerance from $10^{-9}$ to $10^{-10}$.
The final figures for ridge regression include these reruns.

\paragraph{Elastic net and multinomial logistic outcomes.}
The only run that reached its tolerance while failing the common checks was direct SCS on GPU for yearpredictionmsd-rf.
At native tolerance $10^{-7}$ it terminated after 350 iterations in 365.56 seconds, with stationarity $1.178\times10^{-2}$ and feasibility violation $5.002\times10^{-6}$.
We subsequently tried tolerances $10^{-8}$ and $10^{-9}$, but both runs exhausted their 3600-second budgets without a qualifying solution.
Therefore, this entry is listed as a timeout in the final figure.

\subsection{Hardware, Timing, and Solver Settings}
CPU jobs run on nodes with two AMD EPYC 7763 processors (64 physical cores per socket); GPU jobs run on a node with two AMD EPYC 9565 processors (72 physical cores per socket) and NVIDIA H200 NVL GPUs.
Each task receives 64 physical CPU cores and 128 GiB host memory; GPU tasks also receive one 141 GB GPU.

\paragraph{Ridge timing.}
Data generation, initial device placement, and a single warmup are outside the solve timer.
The warmup for iterative methods is capped at ten iterations; QR performs a full warmup solve.
There is one measured solve for each of the three seeds, with a 900-second solve limit, a separate 900-second startup limit, and a total iteration ceiling of $2p$.
GPU timing synchronizes device work.
Warmup failures are recorded separately from measured-solve failures; in particular, the CPU QR failures at the largest square shape occur during warmup.

\paragraph{Bounded elastic net and multinomial logistic timing.}
Each result is one cold solve, without warmup, with a 3600-second solve limit, a separate 1800-second startup limit, and at most 100000 iterations.
Loading, preprocessing, random feature construction, and conversion to each solver's input representation occur outside the timer.
Native solver setup, factorization, JIT compilation (when enabled), and transfers internal to a solver call are included.
\rlaopt{} uses minibatches of size 256 for multinomial logistic regression.
SCS is built with CPU/GPU direct/indirect backends; GPU direct SCS uses cuDSS.
Clarabel uses QDLDL on CPU and cuClarabel uses cuDSS on GPU.
The conic solvers use an equivalent formulation of \eqref{eq:experiment-bounded-en} which avoids forming a dense Gram matrix:
\begin{equation}
    \begin{array}{ll}
        \underset{r\in\R^n,\,w\in\R^p,\,b\in\R}{\text{minimize}}
        & \displaystyle \frac{1}{2n}\|r\|_2^2
          + \frac{\lambda_2}{2}\|w\|_2^2
          + \lambda_1\mathbf1^T w \\[1ex]
        \text{subject to} & r=Xw+b\mathbf1-y, \\
                          & 0\leq w\leq1.
    \end{array}
    \label{eq:experiment-conic-en}
\end{equation}

\paragraph{Failure categories.}
Timeouts, iteration limits, host-memory failures, GPU-memory failures, index overflow, and worker errors remain distinct in the figures and outcome tables.
For the two CPU direct SCS worker errors on acsincome-rf and yearpredictionmsd-rf, the worker connection closed without a terminal solver result.
Despite investigation, we could not determine the exact cause of these failures, but there is a good chance of a memory-related issue, since Clarabel also had memory issues on these two datasets.

\begin{table}[t]
\centering\small
\caption{Software versions used for the experiments.}
\label{tab:experiment-versions}
\begin{tabular}{ll}
\toprule
Software & Version \\
\midrule
\rlaopt{} & 0.1.0 \\
PyTorch & 2.13.0+cu130 \\
NumPy / SciPy & 2.4.2 / 1.18.1 \\
cuML & 26.8.0 \\
JAXopt & 0.8.5 \\
JAX & 0.11.1 \\
SCS (all four backends) & 3.2.11 \\
Julia & 1.10.12 \\
Clarabel / cuClarabel source project & 0.11.0 (pinned revision below) \\
CUDA.jl / CUDSS.jl & 5.11.3 / 0.6.5 \\
cuDSS & 0.7.1 \\
CUDA container base & 13.0.2 \\
\bottomrule
\end{tabular}
\end{table}

\cref{tab:experiment-versions} records the software environment.
Clarabel and cuClarabel use source revision
\texttt{ffa325c89fa90b7e86b745}\allowbreak\texttt{fa61b1dca64daf3a06}.

\paragraph{Differentiable optimization.}
\label{sec:diff-reproduction}
The differentiable optimization experiment uses $2^9$ training and $2^7$ validation observations with $2^6$ standard Gaussian features, $2^4$ nonzero coefficients drawn from $\mathcal N(0,1/2^4)$, and Gaussian noise with standard deviation 0.1.
Each inner solve starts at zero and uses 100 proximal gradient steps with step size $n_{\rm train}/(2\|X_{\rm train}\|_2^2)$.
We take 40 outer steps of size 0.05 starting from $\mu=0.2$.

\subsection{Modeling Examples}
These examples show how to define each problem, initialize the solver, and run a fixed number of iterations.
The ridge operator applies $X^T X$ without materializing it:
\begin{CodeBlock}
\PYG{n}{x\PYGZus{}op} \PYG{o}{=} \PYG{n}{aslinearoperator}\PYG{p}{(}\PYG{n}{X}\PYG{p}{)}
\PYG{n}{normal\PYGZus{}op} \PYG{o}{=} \PYG{n}{x\PYGZus{}op}\PYG{o}{.}\PYG{n}{T} \PYG{o}{@} \PYG{n}{x\PYGZus{}op}
\PYG{n}{lin\PYGZus{}sys} \PYG{o}{=} \PYG{n}{LinSys}\PYG{p}{(}\PYG{n}{normal\PYGZus{}op}\PYG{p}{,} \PYG{p}{(}\PYG{n}{X}\PYG{o}{.}\PYG{n}{T} \PYG{o}{@} \PYG{n}{y}\PYG{p}{)}\PYG{o}{.}\PYG{n}{unsqueeze}\PYG{p}{(}\PYG{o}{\PYGZhy{}}\PYG{l+m+mi}{1}\PYG{p}{)}\PYG{p}{,} \PYG{n}{reg}\PYG{o}{=}\PYG{n}{reg}\PYG{p}{)}
\PYG{n}{precond} \PYG{o}{=} \PYG{n}{NystromConfig}\PYG{p}{(}\PYG{n}{rank\PYGZus{}init}\PYG{o}{=}\PYG{l+m+mi}{128}\PYG{p}{,} \PYG{n}{rank\PYGZus{}max}\PYG{o}{=}\PYG{l+m+mi}{128}\PYG{p}{,}
                       \PYG{n}{base\PYGZus{}damping}\PYG{o}{=}\PYG{n}{reg}\PYG{p}{,} \PYG{n}{damping\PYGZus{}mode}\PYG{o}{=}\PYG{l+s+s2}{\PYGZdq{}}\PYG{l+s+s2}{adaptive}\PYG{l+s+s2}{\PYGZdq{}}\PYG{p}{)}
\PYG{n}{solver} \PYG{o}{=} \PYG{n}{PCG}\PYG{p}{(}\PYG{n}{lin\PYGZus{}sys}\PYG{p}{,} \PYG{n}{PCGConfig}\PYG{p}{(}\PYG{n}{preconditioner\PYGZus{}config}\PYG{o}{=}\PYG{n}{precond}\PYG{p}{)}\PYG{p}{)}

\PYG{n}{params} \PYG{o}{=} \PYG{n}{lin\PYGZus{}sys}\PYG{o}{.}\PYG{n}{w}
\PYG{n}{state} \PYG{o}{=} \PYG{n}{solver}\PYG{o}{.}\PYG{n}{init\PYGZus{}state}\PYG{p}{(}\PYG{n}{params}\PYG{p}{)}
\PYG{k}{for} \PYG{n}{\PYGZus{}} \PYG{o+ow}{in} \PYG{n+nb}{range}\PYG{p}{(}\PYG{l+m+mi}{100}\PYG{p}{)}\PYG{p}{:}
    \PYG{n}{params}\PYG{p}{,} \PYG{n}{state} \PYG{o}{=} \PYG{n}{solver}\PYG{o}{.}\PYG{n}{step}\PYG{p}{(}\PYG{n}{params}\PYG{p}{,} \PYG{n}{state}\PYG{p}{)}
\end{CodeBlock}

For bounded multinomial logistic regression, the coefficient matrix has one column per class:
\begin{CodeBlock}
\PYG{n}{beta} \PYG{o}{=} \PYG{n}{Variable}\PYG{p}{(}\PYG{p}{(}\PYG{n}{X}\PYG{o}{.}\PYG{n}{shape}\PYG{p}{[}\PYG{l+m+mi}{1}\PYG{p}{]}\PYG{p}{,} \PYG{n}{K}\PYG{p}{)}\PYG{p}{,} \PYG{n}{dtype}\PYG{o}{=}\PYG{n}{torch}\PYG{o}{.}\PYG{n}{float64}\PYG{p}{,} \PYG{n}{device}\PYG{o}{=}\PYG{n}{device}\PYG{p}{)}
\PYG{n}{loader} \PYG{o}{=} \PYG{n}{DataLoader}\PYG{p}{(}\PYG{n}{Dataset}\PYG{p}{(}\PYG{n}{X}\PYG{p}{,} \PYG{n}{y}\PYG{p}{,} \PYG{n}{device}\PYG{o}{=}\PYG{n}{device}\PYG{p}{)}\PYG{p}{,} \PYG{n}{batch\PYGZus{}size}\PYG{o}{=}\PYG{l+m+mi}{256}\PYG{p}{)}
\PYG{n}{model} \PYG{o}{=} \PYG{n}{MultinomialRegression}\PYG{p}{(}\PYG{n}{beta}\PYG{p}{,} \PYG{n}{loader}\PYG{p}{,} \PYG{n}{fit\PYGZus{}intercept}\PYG{o}{=}\PYG{k+kc}{False}\PYG{p}{)}
\PYG{n}{obj} \PYG{o}{=} \PYG{n}{model} \PYG{o}{+} \PYG{n}{Box}\PYG{p}{(}\PYG{n}{beta}\PYG{p}{,} \PYG{n}{lower}\PYG{o}{=}\PYG{o}{\PYGZhy{}}\PYG{l+m+mf}{1.0}\PYG{p}{,} \PYG{n}{upper}\PYG{o}{=}\PYG{l+m+mf}{1.0}\PYG{p}{)}
\PYG{n}{solver} \PYG{o}{=} \PYG{n}{Sapphire}\PYG{p}{(}\PYG{n}{obj}\PYG{p}{,} \PYG{n}{config}\PYG{o}{=}\PYG{n}{SapphireConfig}\PYG{p}{(}\PYG{p}{)}\PYG{p}{)}

\PYG{n}{variable\PYGZus{}values} \PYG{o}{=} \PYG{n}{obj}\PYG{o}{.}\PYG{n}{variable\PYGZus{}values}
\PYG{n}{state} \PYG{o}{=} \PYG{n}{solver}\PYG{o}{.}\PYG{n}{init\PYGZus{}state}\PYG{p}{(}\PYG{n}{variable\PYGZus{}values}\PYG{p}{)}
\PYG{k}{for} \PYG{n}{\PYGZus{}} \PYG{o+ow}{in} \PYG{n+nb}{range}\PYG{p}{(}\PYG{l+m+mi}{200}\PYG{p}{)}\PYG{p}{:}
    \PYG{n}{variable\PYGZus{}values}\PYG{p}{,} \PYG{n}{state} \PYG{o}{=} \PYG{n}{solver}\PYG{o}{.}\PYG{n}{step}\PYG{p}{(}\PYG{n}{variable\PYGZus{}values}\PYG{p}{,} \PYG{n}{state}\PYG{p}{)}
\end{CodeBlock}

For bounded elastic net, the model has an unregularized intercept:
\begin{CodeBlock}
\PYG{n}{w} \PYG{o}{=} \PYG{n}{Variable}\PYG{p}{(}\PYG{p}{(}\PYG{n}{X}\PYG{o}{.}\PYG{n}{shape}\PYG{p}{[}\PYG{l+m+mi}{1}\PYG{p}{]}\PYG{p}{,}\PYG{p}{)}\PYG{p}{,} \PYG{n}{dtype}\PYG{o}{=}\PYG{n}{torch}\PYG{o}{.}\PYG{n}{float64}\PYG{p}{,} \PYG{n}{device}\PYG{o}{=}\PYG{n}{device}\PYG{p}{)}
\PYG{n}{loader} \PYG{o}{=} \PYG{n}{DataLoader}\PYG{p}{(}\PYG{n}{Dataset}\PYG{p}{(}\PYG{n}{X}\PYG{p}{,} \PYG{n}{y}\PYG{p}{,} \PYG{n}{device}\PYG{o}{=}\PYG{n}{device}\PYG{p}{)}\PYG{p}{,} \PYG{n}{batch\PYGZus{}size}\PYG{o}{=}\PYG{l+m+mi}{256}\PYG{p}{)}
\PYG{n}{model} \PYG{o}{=} \PYG{n}{LinearRegression}\PYG{p}{(}\PYG{n}{w}\PYG{p}{,} \PYG{n}{loader}\PYG{p}{,} \PYG{n}{fit\PYGZus{}intercept}\PYG{o}{=}\PYG{k+kc}{True}\PYG{p}{)}
\PYG{n}{lambd} \PYG{o}{=} \PYG{l+m+mf}{0.1} \PYG{o}{*} \PYG{n}{torch}\PYG{o}{.}\PYG{n}{linalg}\PYG{o}{.}\PYG{n}{vector\PYGZus{}norm}\PYG{p}{(}\PYG{n}{X}\PYG{o}{.}\PYG{n}{T} \PYG{o}{@} \PYG{p}{(}\PYG{n}{y} \PYG{o}{\PYGZhy{}} \PYG{n}{y}\PYG{o}{.}\PYG{n}{mean}\PYG{p}{(}\PYG{p}{)}\PYG{p}{)}\PYG{p}{,}
                                      \PYG{n+nb}{ord}\PYG{o}{=}\PYG{n+nb}{float}\PYG{p}{(}\PYG{l+s+s2}{\PYGZdq{}}\PYG{l+s+s2}{inf}\PYG{l+s+s2}{\PYGZdq{}}\PYG{p}{)}\PYG{p}{)} \PYG{o}{/} \PYG{n}{X}\PYG{o}{.}\PYG{n}{shape}\PYG{p}{[}\PYG{l+m+mi}{0}\PYG{p}{]}
\PYG{n}{obj} \PYG{o}{=} \PYG{l+m+mf}{0.5} \PYG{o}{*} \PYG{n}{model} \PYG{o}{+} \PYG{n}{ElasticNet}\PYG{p}{(}\PYG{n}{w}\PYG{p}{,} \PYG{n}{l1\PYGZus{}scaling}\PYG{o}{=}\PYG{n}{lambd}\PYG{p}{,}
                              \PYG{n}{l2\PYGZus{}scaling}\PYG{o}{=}\PYG{n}{lambd}\PYG{p}{)} \PYG{o}{+} \PYG{n}{Box}\PYG{p}{(}\PYG{n}{w}\PYG{p}{,} \PYG{l+m+mf}{0.0}\PYG{p}{,} \PYG{l+m+mf}{1.0}\PYG{p}{)}
\PYG{n}{solver} \PYG{o}{=} \PYG{n}{ADMM}\PYG{p}{(}\PYG{n}{obj}\PYG{p}{,} \PYG{n}{config}\PYG{o}{=}\PYG{n}{ADMMConfig}\PYG{p}{(}\PYG{p}{)}\PYG{p}{)}

\PYG{n}{variable\PYGZus{}values} \PYG{o}{=} \PYG{n}{obj}\PYG{o}{.}\PYG{n}{variable\PYGZus{}values}
\PYG{n}{state} \PYG{o}{=} \PYG{n}{solver}\PYG{o}{.}\PYG{n}{init\PYGZus{}state}\PYG{p}{(}\PYG{n}{variable\PYGZus{}values}\PYG{p}{)}
\PYG{k}{for} \PYG{n}{\PYGZus{}} \PYG{o+ow}{in} \PYG{n+nb}{range}\PYG{p}{(}\PYG{l+m+mi}{20}\PYG{p}{)}\PYG{p}{:}
    \PYG{n}{variable\PYGZus{}values}\PYG{p}{,} \PYG{n}{state} \PYG{o}{=} \PYG{n}{solver}\PYG{o}{.}\PYG{n}{step}\PYG{p}{(}\PYG{n}{variable\PYGZus{}values}\PYG{p}{,} \PYG{n}{state}\PYG{p}{)}
\end{CodeBlock}

\section{Additional Experimental Results}
\label{sec:additional-experiments}
\begin{figure}[p]
\centering\includegraphics[width=\linewidth,height=0.39\textheight,keepaspectratio]{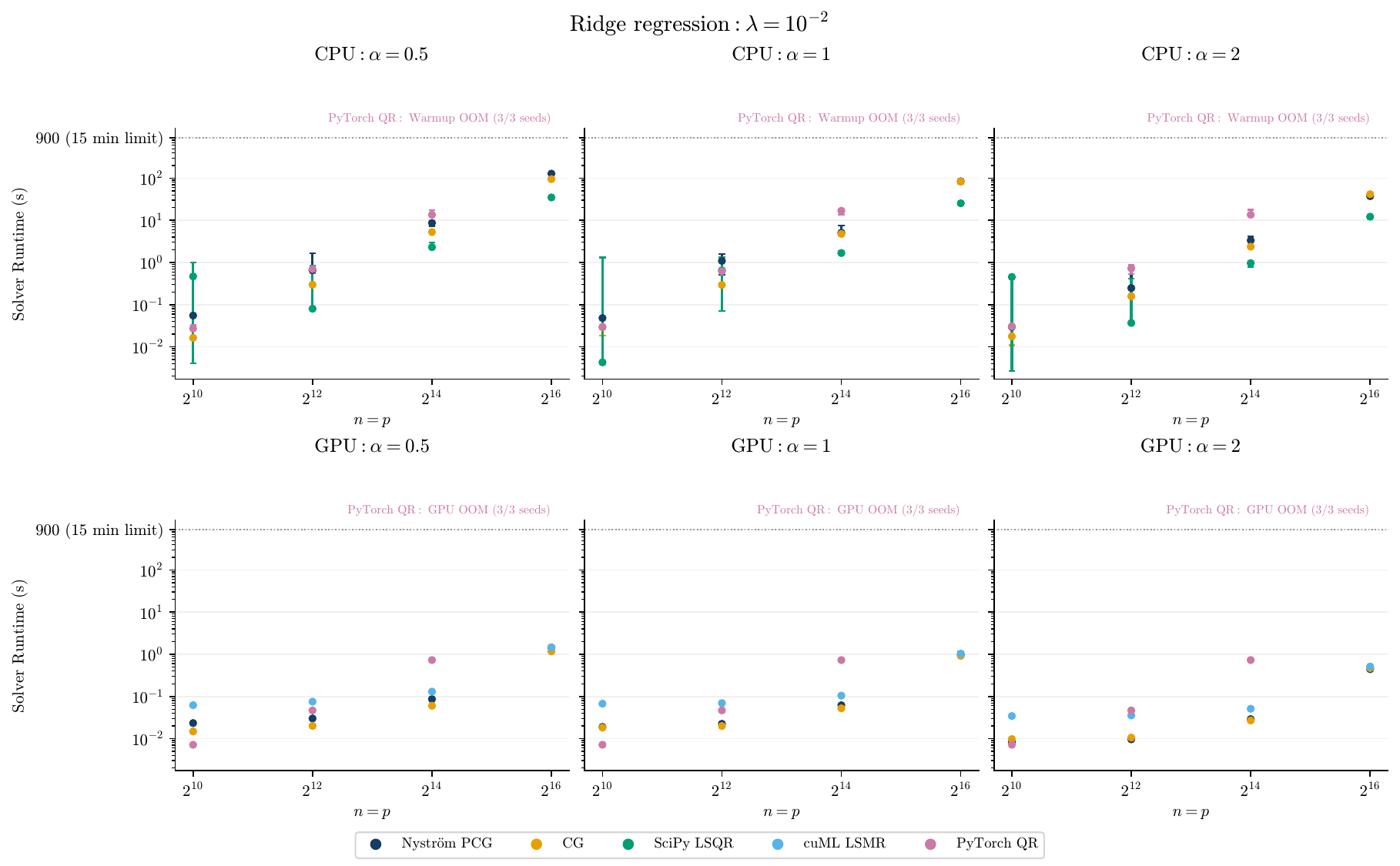}
\caption{Square ridge problems, varying $n=p$, at $\lambda=10^{-2}$. Each point on the plot indicates median solve time over three seeds; the error bars provide the min-max
range of solve times.}
\label{fig:ridge-square-1e-2}
\end{figure}

\begin{figure}[p]
\centering\includegraphics[width=\linewidth,height=0.39\textheight,keepaspectratio]{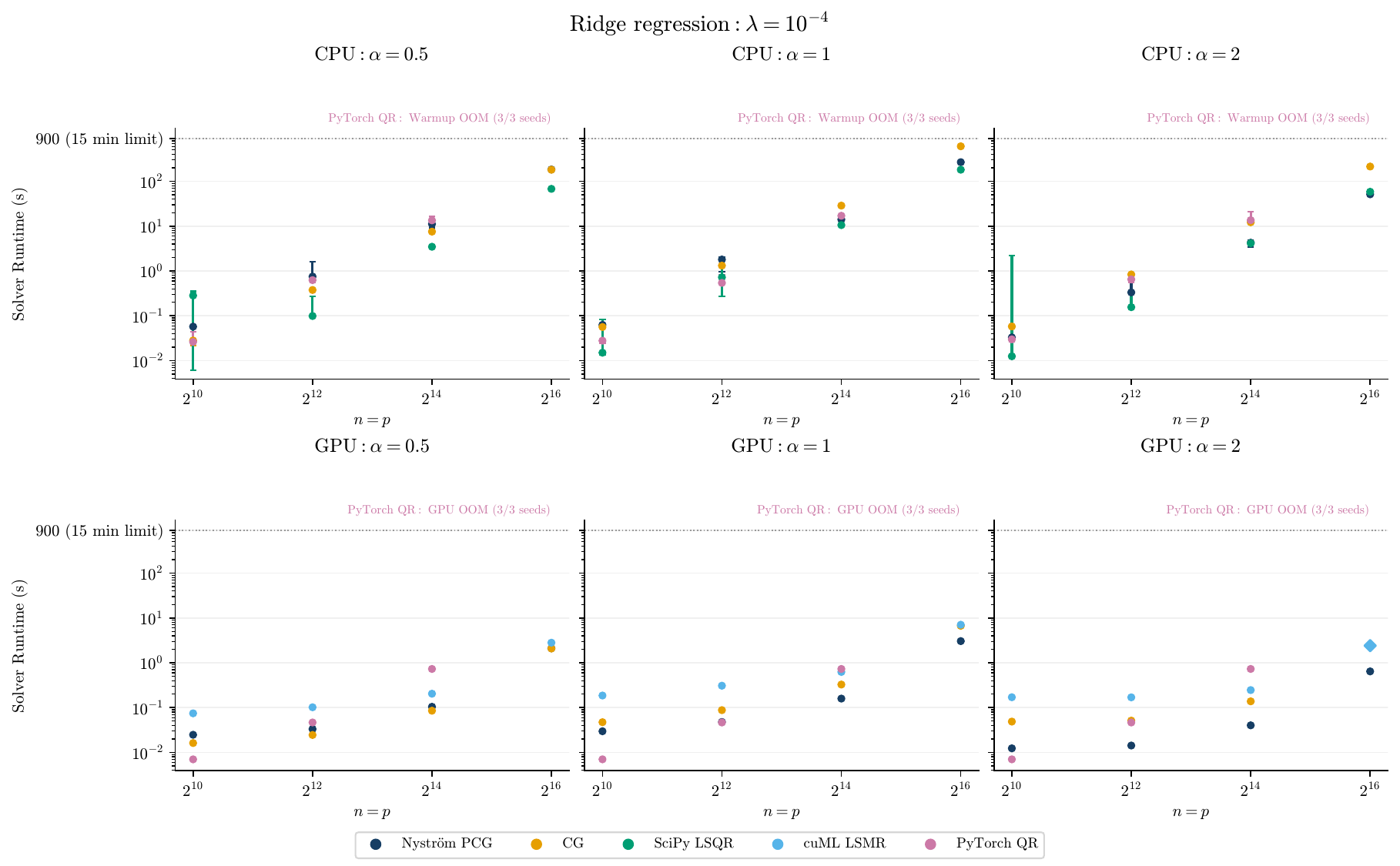}
\caption{Square ridge problems, varying $n=p$, at $\lambda=10^{-4}$. Each point on the plot indicates median solve time over three seeds; the error bars provide the min-max
range of solve times.}
\label{fig:ridge-square-1e-4}
\end{figure}

\begin{figure}[p]
\centering\includegraphics[width=\linewidth,height=0.39\textheight,keepaspectratio]{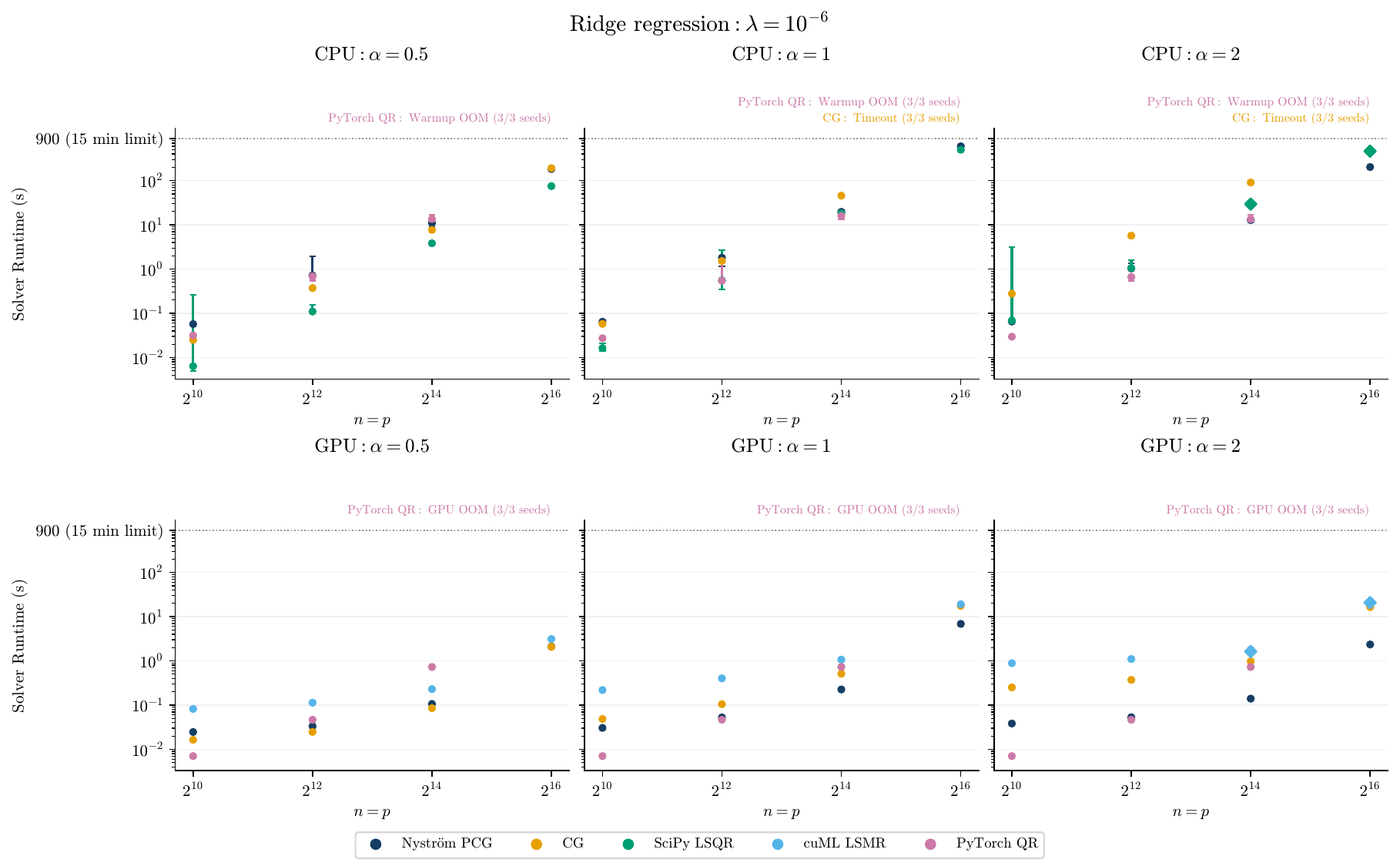}
\caption{Square ridge problems, varying $n=p$, at $\lambda=10^{-6}$. Each point on the plot indicates median solve time over three seeds; the error bars provide the min-max
range of solve times.}
\label{fig:ridge-square-1e-6}
\end{figure}

\begin{figure}[p]
\centering\includegraphics[width=\linewidth,height=0.39\textheight,keepaspectratio]{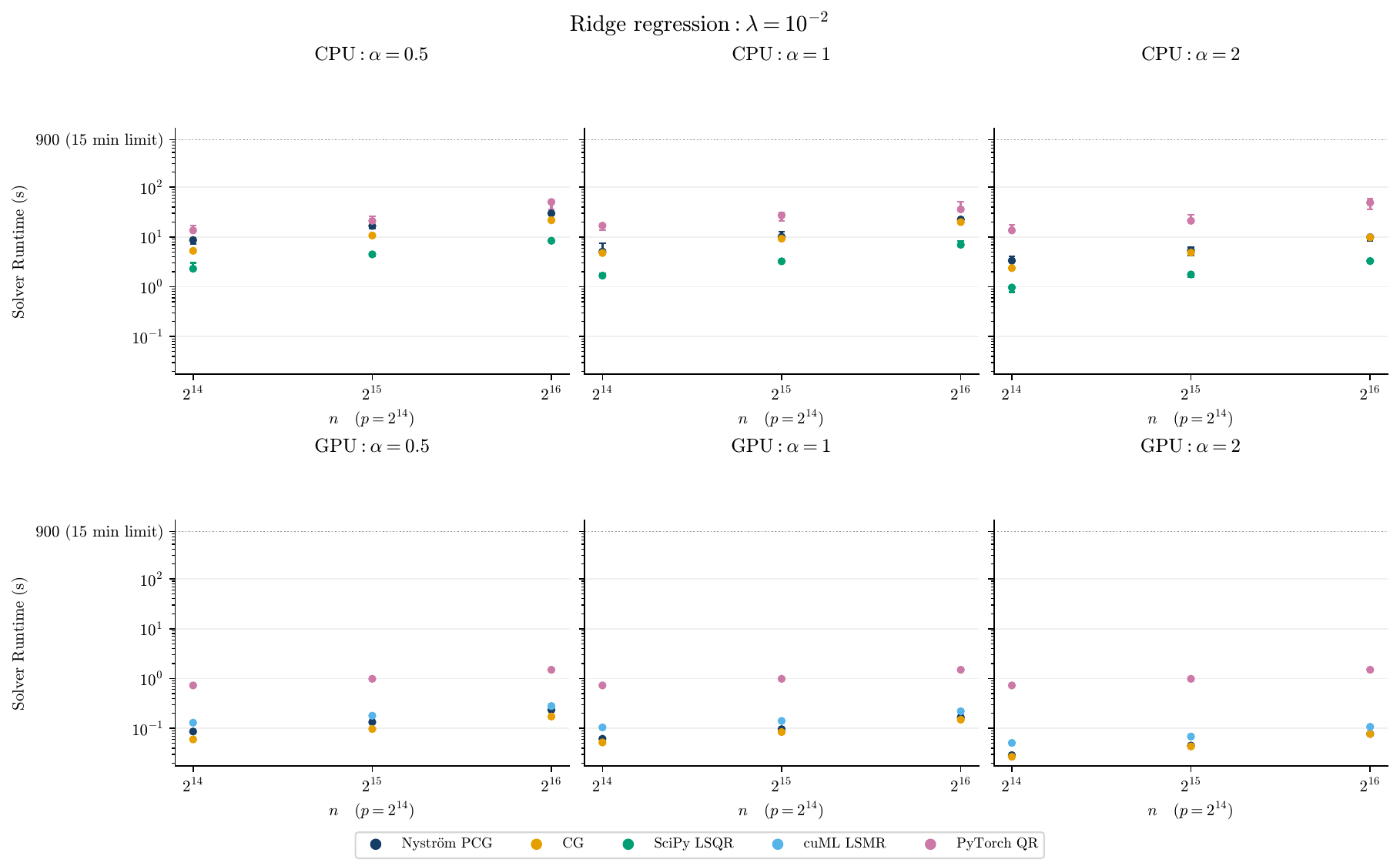}
\caption{Ridge problems with $p=2^{14}$, varying $n$, at $\lambda=10^{-2}$. Each point on the plot indicates median solve time over three seeds; the error bars provide the min-max
range of solve times.}
\label{fig:ridge-fixed_p-1e-2}
\end{figure}

\begin{figure}[p]
\centering\includegraphics[width=\linewidth,height=0.39\textheight,keepaspectratio]{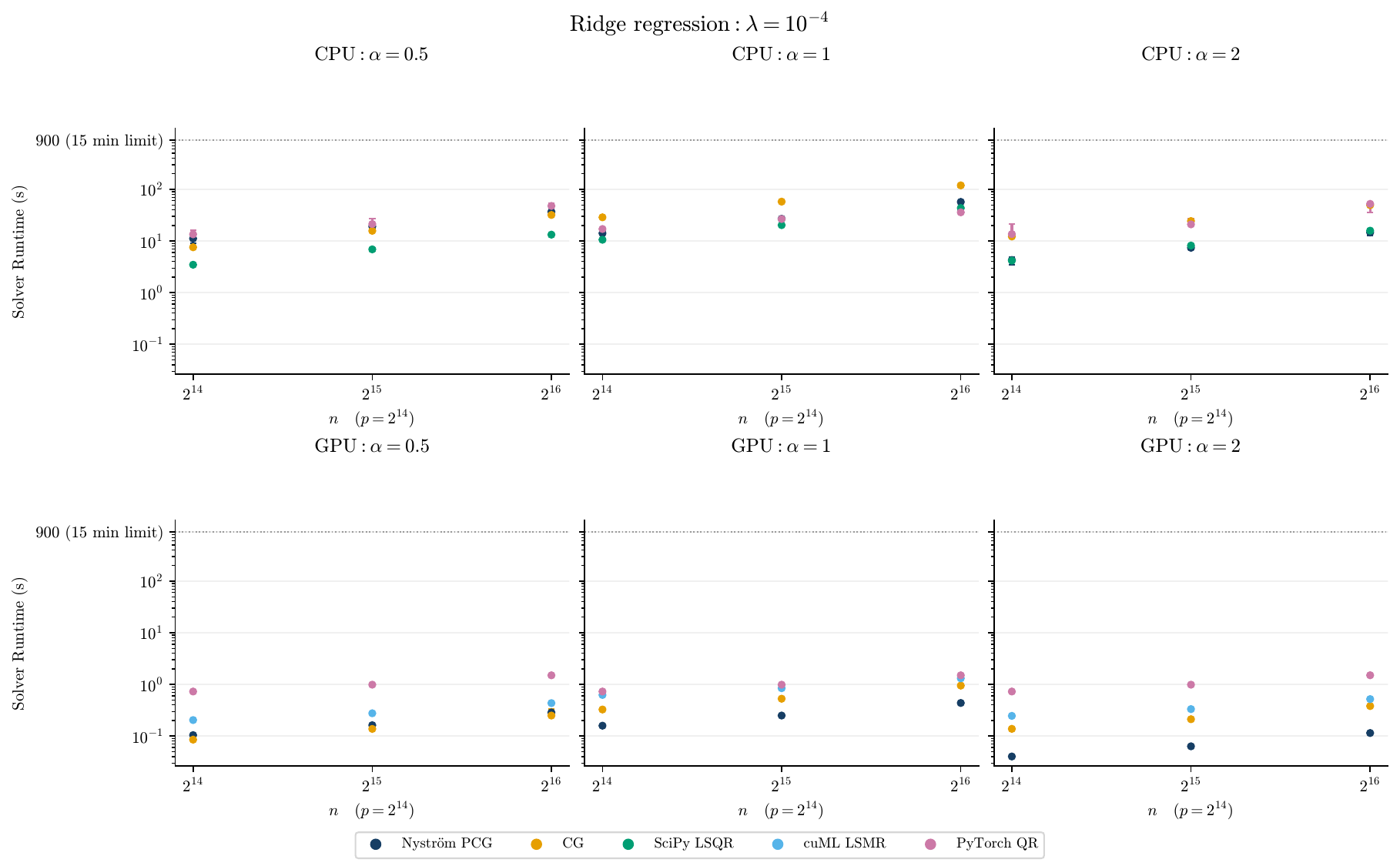}
\caption{Ridge problems with $p=2^{14}$, varying $n$, at $\lambda=10^{-4}$. Each point on the plot indicates median solve time over three seeds; the error bars provide the min-max
range of solve times.}
\label{fig:ridge-fixed_p-1e-4}
\end{figure}

\begin{figure}[p]
\centering\includegraphics[width=\linewidth,height=0.39\textheight,keepaspectratio]{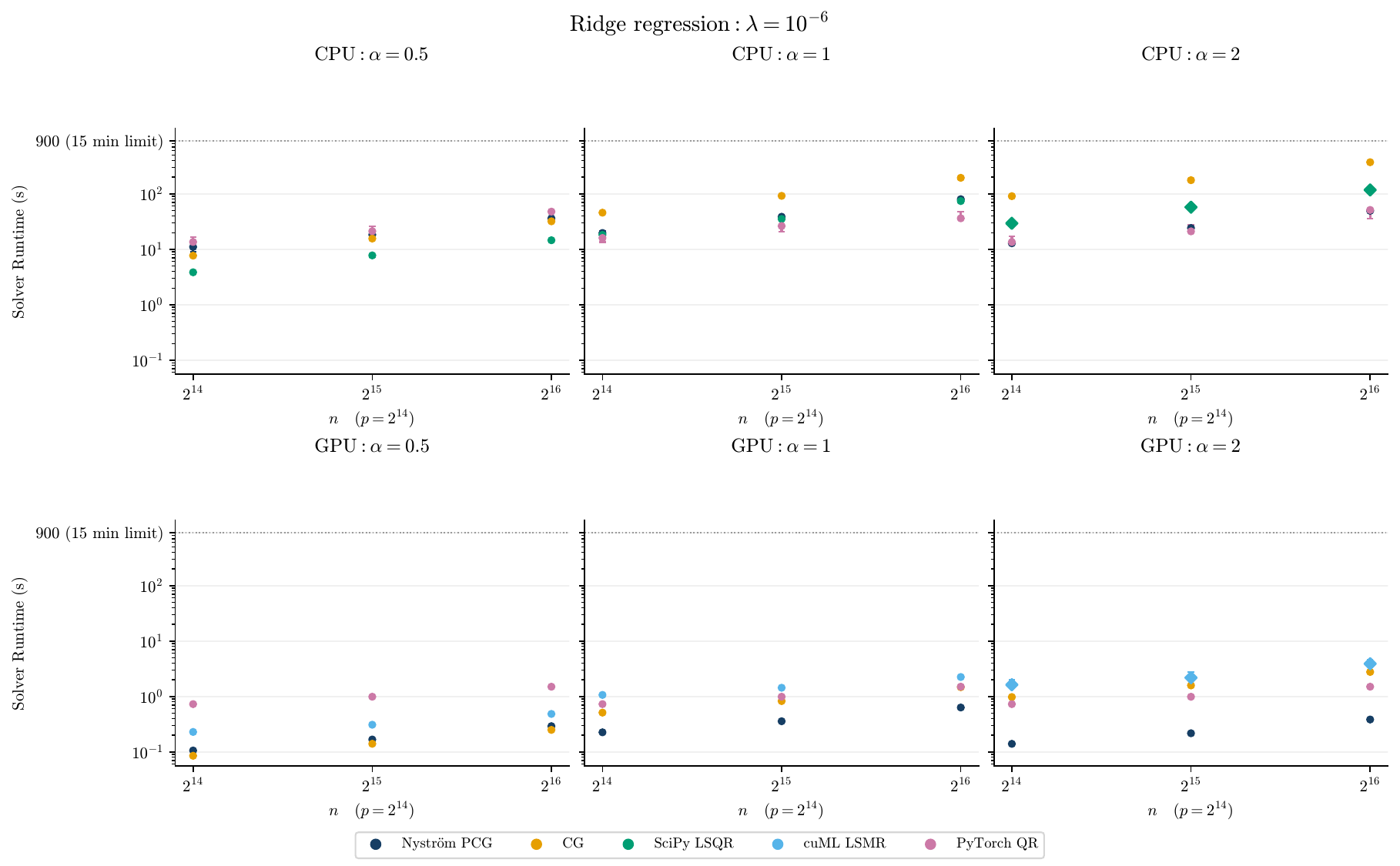}
\caption{Ridge problems with $p=2^{14}$, varying $n$, at $\lambda=10^{-6}$. Each point on the plot indicates median solve time over three seeds; the error bars provide the min-max
range of solve times.}
\label{fig:ridge-fixed_p-1e-6}
\end{figure}

\begin{figure}[p]
\centering\includegraphics[width=\linewidth,height=0.39\textheight,keepaspectratio]{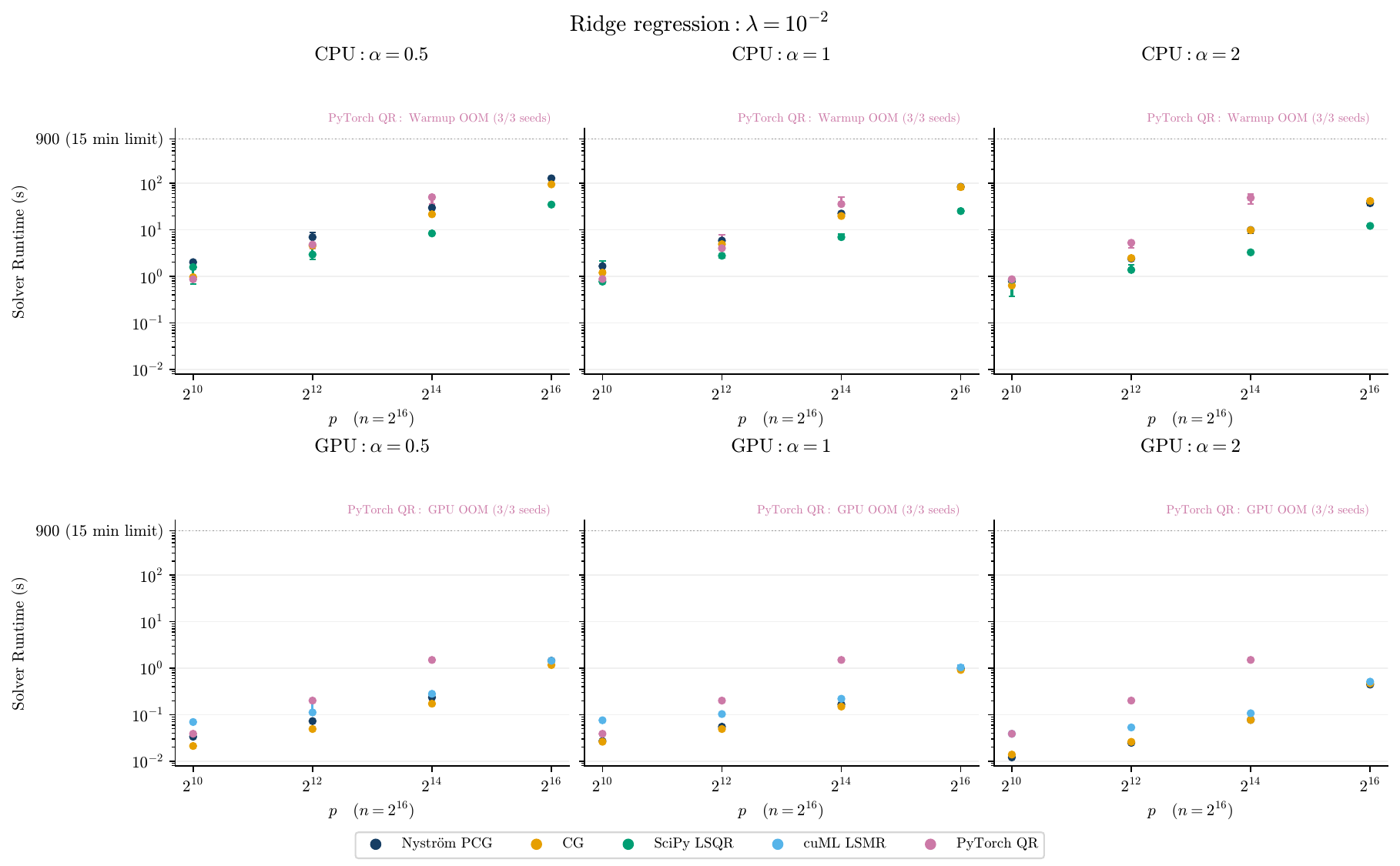}
\caption{Ridge problems with $n=2^{16}$, varying $p$, at $\lambda=10^{-2}$. Each point on the plot indicates median solve time over three seeds; the error bars provide the min-max
range of solve times.}
\label{fig:ridge-fixed_n-1e-2}
\end{figure}

\begin{figure}[p]
\centering\includegraphics[width=\linewidth,height=0.39\textheight,keepaspectratio]{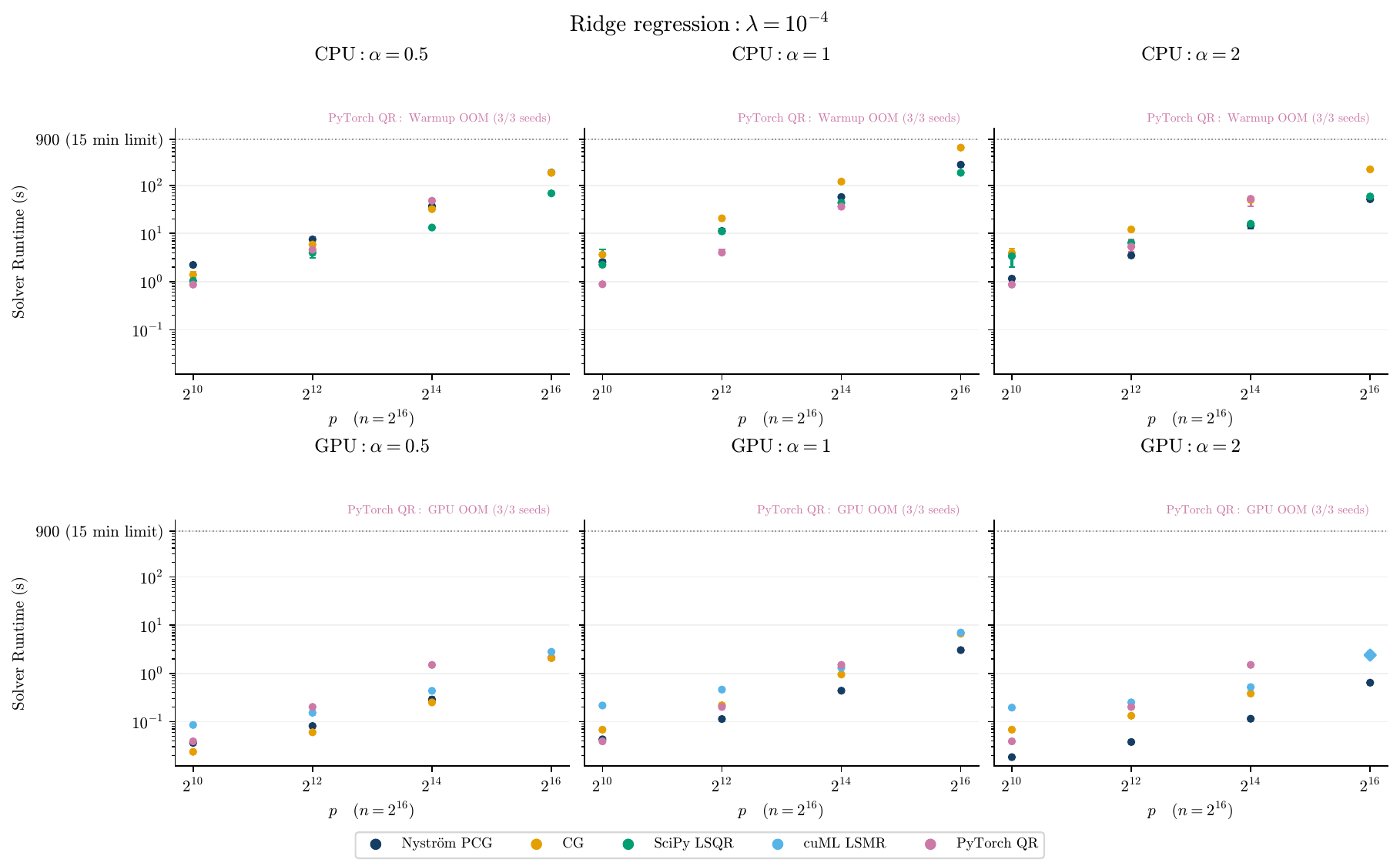}
\caption{Ridge problems with $n=2^{16}$, varying $p$, at $\lambda=10^{-4}$. Each point on the plot indicates median solve time over three seeds; the error bars provide the min-max
range of solve times.}
\label{fig:ridge-fixed_n-1e-4}
\end{figure}

\begin{figure}[p]
\centering\includegraphics[width=\linewidth,height=0.85\textheight,keepaspectratio]{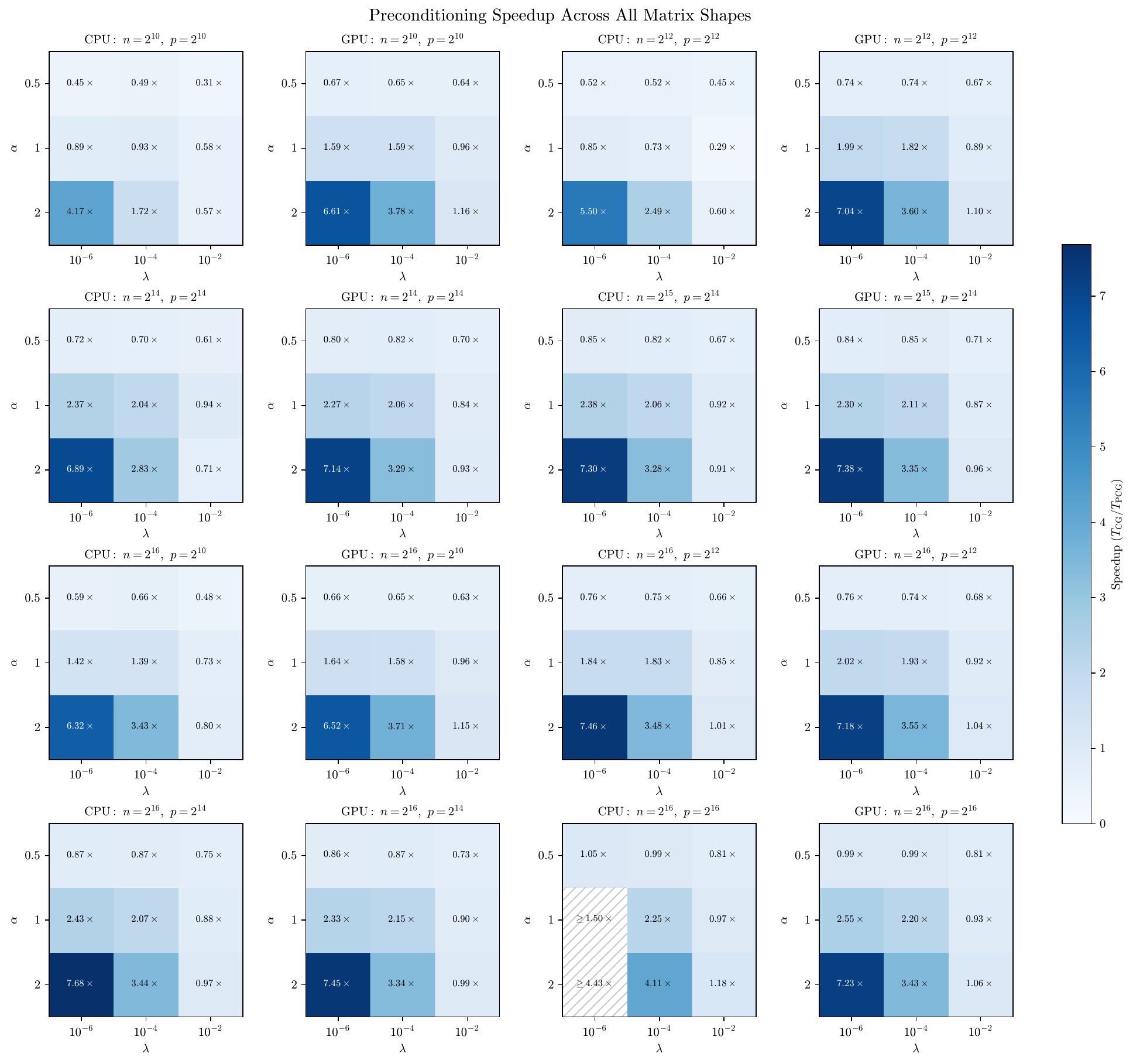}
\caption{CG solve time divided by \nys{}PCG solve time across all eight shapes of $X$, spectral decay rates $\alpha$, and regularization levels $\lambda$. Inequalities denote lower bounds when CG times out.}
\label{fig:ridge-all-preconditioning}
\end{figure}

\begin{figure}[p]
\centering\includegraphics[width=\linewidth]{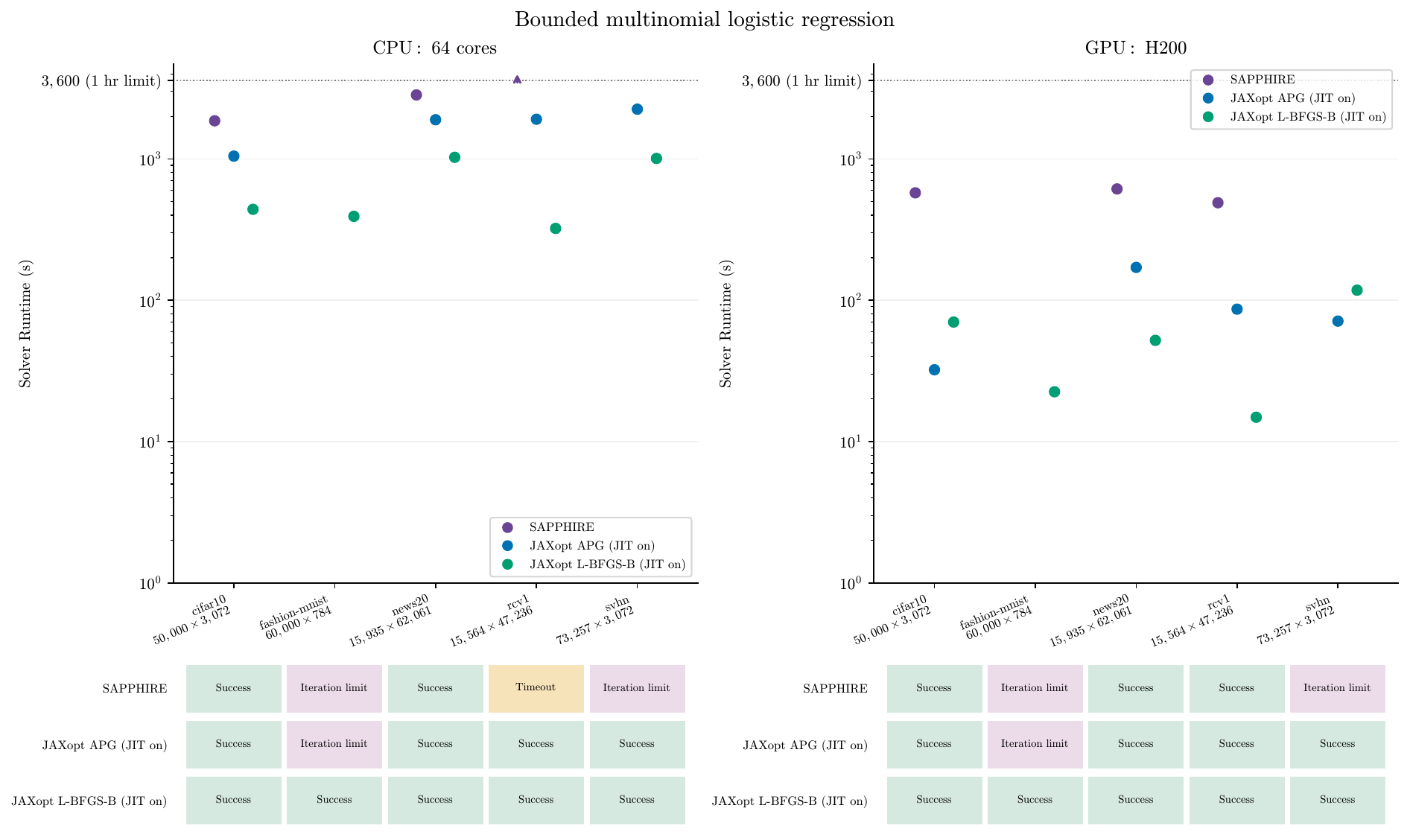}
\caption{Bounded multinomial logistic regression with JAXopt JIT compilation enabled and included in the solve time. The JIT-enabled baselines are faster than SAPPHIRE by at least one order of magnitude.}
\label{fig:multinomial-jit}
\end{figure}

\cref{fig:ridge-square-1e-2,fig:ridge-square-1e-4,fig:ridge-square-1e-6} show the square ridge problems across all three regularization levels.
\cref{fig:ridge-fixed_p-1e-2,fig:ridge-fixed_p-1e-4,fig:ridge-fixed_p-1e-6} and \cref{fig:ridge-fixed_n-1e-2,fig:ridge-fixed_n-1e-4} complete the fixed-$p$ and fixed-$n$ sweeps, respectively.
\cref{fig:ridge-all-preconditioning} summarizes the effects of preconditioning across all eight shapes, and \cref{fig:multinomial-jit} provides the JIT-enabled multinomial logistic regression comparison.
They use the same common accuracy checks and resource limits as the figures in the main paper.

\end{document}